\documentclass{article}

\usepackage[nonatbib, final]{neurips_2022}

\usepackage[utf8]{inputenc} 
\usepackage[T1]{fontenc}    
\usepackage{url}            
\usepackage{booktabs}       
\usepackage{amsfonts}       
\usepackage{nicefrac}       
\usepackage{microtype}      
\usepackage{xcolor}         

\usepackage{amssymb,amsmath}
\usepackage{algorithm}
\usepackage{algorithmicx,algpseudocode}
\usepackage[autostyle]{csquotes}
\usepackage[capitalise]{cleveref}
\usepackage{caption,subcaption}
\usepackage{wrapfig}
\usepackage{amsthm,epsfig}
\usepackage{multirow}
\usepackage{booktabs} 

\newtheorem{theorem}{Theorem}

\newtheorem{definition}{Definition}
\usepackage{thmtools,thm-restate}

\expandafter\def\expandafter\normalsize\expandafter{%
    \normalsize
    \setlength\abovedisplayskip{2pt}
    \setlength\belowdisplayskip{2pt}
    \setlength\abovedisplayshortskip{2pt}
    \setlength\belowdisplayshortskip{2pt}
}

\usepackage{amsmath,amsfonts,bm}

\def\eqref#1{equation~\ref{#1}}

\def\1{\bm{1}}

\def\vg{{\bm{g}}}

\def\vw{{\bm{w}}}
\def\vx{{\bm{x}}}
\def\vy{{\bm{y}}}

\DeclareMathAlphabet{\mathsfit}{\encodingdefault}{\sfdefault}{m}{sl}
\SetMathAlphabet{\mathsfit}{bold}{\encodingdefault}{\sfdefault}{bx}{n}

\def\gD{{\mathcal{D}}}

\def\gF{{\mathcal{F}}}

\def\gL{{\mathcal{L}}}

\def\sD{{\mathbb{D}}}

\def\sP{{\mathbb{P}}}

\def\sT{{\mathbb{T}}}

\newcommand{\proj}{\mathrm{Proj}}

\DeclareMathOperator*{\argmin}{arg\,min}

\title{Beyond Not-Forgetting: Continual Learning with Backward Knowledge Transfer}

\author{%
  Sen Lin \\
  School of ECEE\\
  Arizona State University\\
  \texttt{slin70@asu.edu} \\
   \And
   Li Yang \\
   School of ECEE \\
   Arizona State University \\
   \texttt{lyang166@asu.edu} \\
   \AND
   Deliang Fan \\
   School of ECEE \\
   Arizona State University \\
   \texttt{dfan@asu.edu} \\
   \And
   Junshan Zhang \\
   Department of ECE \\
   University of California, Davis \\
   \texttt{jazh@ucdavis.edu} \\
}

\begin{document}

\maketitle

\begin{abstract}

By learning a sequence of tasks continually, an agent in continual learning (CL) can improve the learning performance of both a new task and `old' tasks by leveraging the forward knowledge transfer and the backward knowledge transfer, respectively. However, most existing CL methods focus on addressing catastrophic forgetting in neural networks by minimizing the modification of the learnt model for old tasks. This inevitably limits the backward knowledge transfer from the new task to the old tasks, because judicious model updates could possibly improve the learning performance of the old tasks as well. To tackle this problem, we first theoretically analyze  the conditions under which updating the learnt model of old tasks could be beneficial for CL and also lead to backward knowledge transfer, based on the gradient projection onto the input subspaces of old tasks. Building on the theoretical analysis, we next develop a ContinUal learning method with Backward knowlEdge tRansfer (CUBER), for a fixed capacity neural network without data replay. In particular,
CUBER first characterizes the task correlation to identify the positively correlated old tasks in a layer-wise manner, and then selectively modifies the learnt model of the old tasks when learning the new task. Experimental studies show that CUBER can even achieve positive backward knowledge transfer on several existing CL benchmarks for the first time without data replay, where the related baselines still suffer from catastrophic forgetting (negative backward knowledge transfer). The superior performance of CUBER on the backward knowledge transfer also leads to higher accuracy accordingly.


\end{abstract}

\section{Introduction}

One ultimate goal of artificial intelligence is to build an agent that can continually learn a sequence of different tasks, so as to echo the remarkable learning capability of human beings during their lifespan. With more tasks being learnt, the agent is expected to be able to learn a new task more easily by leveraging the accumulated knowledge from old tasks (forward knowledge transfer), and also further improve the learning performance of old tasks based on the gained knowledge of related new tasks (backward knowledge transfer). Such a learning paradigm is known as continual learning (CL) \cite{chen2018lifelong,ring1994continual}, which has recently  attracted much attention.

Most existing CL methods focus on addressing the catastrophic forgetting problem \cite{mccloskey1989catastrophic}, i.e., the neural network may easily forget the knowledge of old tasks when learning a new task.  
The main strategy therein is to avoid the model change of old tasks when learning the new task. For example, regularization-based methods (e.g., \cite{kirkpatrick2017overcoming, aljundi2018memory, liu2022continual}) penalize the modification of important weights of old tasks; parameter-isolation based methods (e.g., \cite{fernando2017pathnet,serra2018overcoming, yoon2017lifelong, hung2019compacting}) fix the model learnt for old tasks; and memory-based methods (e.g., \cite{chaudhry2018efficient, farajtabar2020orthogonal, saha2021gradient}) aim to update the model with minimal interference introduced to old tasks.
While effectively mitigating catastrophic forgetting, such a model-freezing strategy inevitably limits the backward knowledge transfer, by implicitly and conservatively treating the model update of the new task as interference to old tasks. 
Intuitively, careful modifications of the learnt model of old tasks may further improve the learning performance especially when the new task shares similar knowledge to the old tasks.

This work seeks to explore CL from a new perspective by going beyond merely addressing forgetting: \emph{When and how could we improve the learnt model of old tasks to facilitate backward knowledge transfer?} In particular, we consider a fixed neural network and the scenario where the data of old tasks is not accessible when learning a new task, which naturally rules out the expansion-based methods (e.g., \cite{yoon2017lifelong,hung2019compacting}) and the experience-replay methods (e.g., \cite{riemer2018learning, chaudhry2019continual}). Clearly, achieving backward knowledge transfer is very challenging in this case as mitigating forgetting herein  is already nontrivial. 
To tackle this challenge, note that
the orthogonal-projection based CL methods (e.g., \cite{saha2021gradient,lin2022trgp}), which
update the model with gradients orthogonal to the input subspaces of old tasks, have
demonstrated the state-of-the-art performance under the setting above. Thus motivated, we propose a new orthogonal-projection based CL method to carefully modify the learnt model of old tasks for better backward knowledge transfer.

The main contributions of this work include  1) the quantification of the conditions under which modifying the learnt model of old tasks is beneficial and 2) a new CL method inspired by the analysis, so as to answer the question above. More specifically,
we first introduce notions of `sufficient  projection' and `positive correlation' based on the gradient projection  onto the subspaces of old tasks to characterize the task correlation. We theoretically show that when the task gradients are sufficiently aligned in the old task subspace, appropriately updating the learnt model of old tasks when learning the new task could improve the learning performance and possibly lead to a better model for the old tasks. Based on this analysis, we next propose an orthogonal-projection based ContinUal learning method with Backward knowlEdge tRansfer (CUBER), which 1) first identifies the positively correlated old tasks for the new task and 2) carefully updates the learnt model of selected old tasks along with the  model learning for the new task. 
The experimental results on standard CL benchmarks show that CUBER can achieve the best backward knowledge transfer compared to all the considered baseline CL methods, which consequently improves the overall learning performance. In particular, to the best of our knowledge, CUBER is the first to demonstrate positive backward knowledge transfer on these benchmarks using a fixed capacity network without experience replay, whereas all the compared baselines still suffer from catastrophic forgetting with negative backward knowledge transfer.

\section{Related work}
\label{related}

\textbf{Continual learning.~} Existing CL methods can be generally divided into three categories: 1) \emph{Regularization-based methods} (e.g., \cite{kirkpatrick2017overcoming, aljundi2018memory, lee2017overcoming,liu2022continual}) add additional regularization in the loss function to penalize the model change of the old tasks when learning the new task. For example, Elastic Weight Consolidation (EWC) \cite{kirkpatrick2017overcoming} regularizes the update on the important weights that are evaluated using the Fisher Information matrix; a recursive gradient optimization method is proposed in \cite{liu2022continual} to modify the gradient direction for the objective function regularized by the model change during CL. 2) \emph{Parameter-isolation based methods} (e.g., \cite{rusu2016progressive, li2017learning,yoon2017lifelong, yoon2020scalable, serra2018overcoming, yang2021grown}) allocate different model parameters to different tasks, fix the parameters for old tasks by masking them out during the new task learning and expand the network when needed. For example,   Additive Parameter Decomposition (APD)  \cite{yoon2020scalable} proposes a hierarchical knowledge consolidation method to separate task-specific parameters and task-shared parameters,  where the task-specific parameters are kept intact to address forgetting.
3) \emph{Memory-based methods} can be further divided into experience-replay  methods and orthogonal-projection based methods depending on if the data of old tasks is available for the new task. Experience-replay methods (e.g., \cite{chaudhry2018efficient, riemer2018learning, guo2020improved}) store and replay the old tasks data when learning the new task, while
orthogonal-projection based methods
(e.g., \cite{zeng2019continual,farajtabar2020orthogonal,saha2021gradient,lin2022trgp}) update the model in the orthogonal direction of old tasks without storing the raw data of old tasks. In particular, \cite{kao2021natural} proposes natural continual learning based on Bayesian learning to unify weight regularization and orthogonal gradient projection. However, \cite{kao2021natural} leverages the knowledge of the current task in an implicit way to improve on the old tasks, whereas our method attempts more explicitly to update the model of old tasks whenever relevant new information is available.

\textbf{Knowledge transfer.~} Most of CL methods focus on the catastrophic forgetting problem, and little attention has been put on the knowledge transfer across tasks in CL with neural networks. Progressive Network \cite{rusu2016progressive} considers the forward knowledge transfer but learns one model for each task.
\cite{lin2022trgp} proposes trust region gradient projection (TRGP) to facilitate forward knowledge transfer from correlated old tasks to the new task through a scaling matrix, but the model is still updated along the direction orthogonal to the input subspaces of old tasks to mitigate forgetting. \cite{ke2020continual} studies the knowledge transfer in a mixed sequence of similar and dissimilar tasks, but employs a complicated task similarity detection mechanism where two additional networks need to be trained first for each task before learning the new task.

\section{When could we improve the learnt model of old tasks?}
\label{headings}

In continual learning, a sequence of tasks $\sT=\{t\}_{t=1}^T$ arrives sequentially. For each task $t$, there is a dataset $\sD^t=\{(\vx^t_{i},\vy^t_{i})\}_{i=1}^{N^t}$ with $N^t$ sample pairs,  which is sampled from some unknown distribution $\sP^t$. In this work, we
consider a fixed capacity neural network with weights $\vw$. When learning a new task $t$, we only have access to the dataset $\sD^t$ and no datapoints of old tasks are available. We further denote $\gL(\vw, \sD^t)=\gL_t(\vw)$ as the loss function for training, e.g., mean squared and cross-entropy loss, and $\vw^t$ as the model after learning task $t$.



Intuitively, if the new task $t$ has strong similarity with some old tasks, appropriate model update of the new task would not introduce forgetting of the similar old tasks, but  can lead to a better model for these old tasks due to the backward knowledge transfer. 
To formally characterize this task similarity, we introduce the following conditions on the task gradients.

\begin{definition}[Sufficient Projection]
\label{def:1}
For any new task $t\in [1,T]$, we say it has sufficient gradient projection on the input subspace of old task $j\in [0, t-1]$ if for some $\epsilon_1\in (0, 1)$
{\small
\begin{align*}
    \|\proj_{S^j}(\nabla \gL_t(\vw^{t-1}))\|_2\geq \epsilon_1\|\nabla \gL_t(\vw^{t-1})\|_2.
\end{align*}}%
\end{definition}

Here $\proj_{S^j}$ defines the projection on the input subspace $S^j$ of task $j$: $\proj_{S^j}(A)=AB^j(B^j)'$ for some matrix $A$ and $B^j$ is the bases for $S^j$.
This definition of sufficient projection follows the same line of the trust region defined in \cite{lin2022trgp}, which implies that task $t$ and $j$ may have sufficient common bases between their input subspaces and hence are strongly correlated, because the gradient lies in the span of the input \cite{zhang2021understanding}.

\begin{definition}[Positive Correlation]
\label{def:2}
For any new task $t\in [1,T]$, we say it has positive correlation with old task $j\in [0, t-1]$ if
for some $\epsilon_2\in (0, 1)$
{\small
\begin{align*}
    \langle\nabla \gL_j(\vw^j), \nabla \gL_t(\vw^{t-1}) \rangle\geq \epsilon_2 \|\nabla \gL_j(\vw^j)\|_2\|\nabla \gL_t(\vw^{t-1})\|_2.
\end{align*}}%
\end{definition}

While sufficient projection suggests the possibly strong correlation between two tasks $t$ and $j$, Definition \ref{def:2} goes one step further  by introducing the positive correlation between tasks, in the sense that the initial model gradient of task $t$ is not conflicting with the model gradient when learning task $j$. In fact, it can be shown that the positive correlation implies the sufficient projection in Definition \ref{def:1} for some $\epsilon_1\geq \epsilon_2$. Note that both Definition \ref{def:1} and \ref{def:2} characterize the correlation based on the initial model $\vw^{t-1}$, which allows the task correlation detection before learning the new task $t$.

For ease of exposition, consider the scenario with a sequence of two tasks 1 and 2. Let $\gF(\vw)=\gL(\vw, \gD_1)+\gL(\vw, \gD_2)$, $\vg_1(\vw)=\nabla_{\vw} \gL(\vw, \gD_1)$ and $\vg_2(\vw)=\nabla_{\vw} \gL(\vw, \gD_2)$. Given the model learnt for task 1 as $\vw^1$, we consider the following two model update rules for learning task 2, \emph{where only the data of task 2 is available}:
\begin{itemize}
    \item (Rule $\#1$): $\vw_{k+1}=\vw_k-\alpha[\vg_2(\vw_k)-\proj_{S^1}(\vg_2(\vw_k))]$,
    \item (Rule $\#2$): $\vw_{k+1}=\vw_k-\alpha \vg_2(\vw_k)$.
\end{itemize}
Here $\vw_0=\vw^1$ and $k\in [0,K-1]$.
Rule $\#1$ characterizes the model update of the orthogonal-projection based methods, where no interference is introduced to task 1 as the model change is orthogonal to the input subspace $S^1$ of task 1 \cite{lin2022trgp} (and hence the learnt model of task 1 will not be updated). In contrast, the vanilla gradient descent Rule $\#2$ for learning task 2 will inevitably modify the learnt model $\vw^1$ for task 1 unless $\proj_{S^1}(\vg_2(\vw_k))=0$, i.e., the extreme case where both forgetting and knowledge transfer do not occur. In what follows, we  evaluate the performance of these update rules.

\begin{theorem}
\label{thm:1}
 Suppose that the loss $\gL$ is $B$-Lipschitz and $\frac{H}{2}$-smooth. Let $\alpha < \min\left\{\frac{1}{H},\frac{\gamma \|\vg_1(\vw_0)\|}{HBK}\right\}$ and $\epsilon_2\geq \frac{(2+\gamma^2)\|\vg_1(\vw_0)\|}{4\|\vg_2(\vw_0)\|}$ for some $\gamma\in (0,1)$. We have the following results:
 
(1) If $\gL$ is convex, Rule $\#2$ for task 2  converges to the optimal model $\vw^*=\argmin~\gF(\vw)$;

(2) If $\gL$ is nonconvex, Rule $\#2$ for task 2 converges to the first order stationary point, i.e.,
{\small
\begin{align*}
    \min_{k} \|\nabla \gF(\vw_k)\|^2<\frac{2}{\alpha K}[\gF(\vw_0)-\gF(\vw^*)]+\frac{4+\gamma^2}{2}\|\vg_1(\vw_0)\|^2.
\end{align*}
}%
\end{theorem}

Theorem \ref{thm:1} indicates that updating the model with Rule $\#2$ will lead to the convergence to the minimizer of the joint objective function $\gF(\vw)$ in the convex case, and the convergence to the first order stationary point in the nonconvex case, when task 1 and 2 satisfy the positive correlation with $\epsilon_2\geq \frac{(2+\gamma^2)\|\vg_1(\vw_0)\|}{4\|\vg_2(\vw_0)\|}$. That is to say, Rule $\#2$ not only results in a good model for task 2, but can also be beneficial for the joint learning of task 1 and 2.
Note that since $\vw_0$ is the learnt model of task 1, in general we have $\|\vg_1(\vw_0)\|<\|\vg_2(\vw_0)\|$. Proof of Theorem \ref{thm:1} can be found in Appendix A.

\begin{theorem}
\label{thm:2}
Suppose that the loss $\gL$ is $B$-Lipschitz and $\frac{H}{2}$-smooth. We have the following results:

(1) Let $\vw^c$ and $\vw^r$ be the model parameters after applying one update to some initial model $\vw$ by using Rule $\#1$ and Rule $\#2$, respectively.
Suppose $\alpha < \min\left\{\frac{1}{H},\frac{\gamma \|\vg_1(\vw_0)\|}{HBK}\right\}$, $\epsilon_1\geq \sqrt{\frac{1+2\alpha H}{2+\alpha H}}$ and $\epsilon_2\geq \frac{(2+\gamma^2)\|\vg_1(\vw_0)\|}{4\|\vg_2(\vw_0)\|}$ for some $\gamma\in (0,1)$. It follows that $\gF(\vw^r)\leq \gF(\vw^c)$;

(2) Let $\vw_k$ be the $k$-th iterate for task 2 with Rule $\#2$. Suppose that $\langle \vg_1(\vw_0), \vg_2(\vw_i)\rangle\geq \epsilon_2 \|\vg_1(\vw_0)\|\|\vg_2(\vw_i)\|$ for $i\in [0,k-1]$ and $\alpha\leq \frac{4\epsilon_2\|\vg_1(\vw_0)\|}{HBk^{1.5}}$. It follows that $\gL_1(\vw_k)\leq \gL_1(\vw^1)$.
\end{theorem}

Intuitively, the first part of Theorem \ref{thm:2} shows that updating with Rule $\#2$ can achieve lower loss value compared to Rule $\#1$ after one step gradient update when task 1 and 2 satisfy the sufficient projection with $\epsilon_1\geq\sqrt{\frac{1+2\alpha H}{2+\alpha H}}$ and the positive correlation with $\epsilon_2\geq \frac{(2+\gamma^2)\|\vg_1(\vw_0)\|}{4\|\vg_2(\vw_0)\|}$. If the positive correlation condition also holds for iterates of Rule $\#2$ when learning the task 2, the second part of Theorem \ref{thm:2} indicates that updating the model indeed leads to a better model for task 1 with respect to $\gL_1$. In a nutshell, when 1) the task 2 has sufficient gradient projection onto the subspace of task 1 and 2) the projected gradient is also aligned well with the gradient of task 1 in the subspace of task 1, updating the model along $\vg_2$ will modify the learnt model $\vw^1$ towards a favorable direction for CL and enable the backward knowledge transfer to task 1.  Proof of Theorem \ref{thm:2} is in Appendix B.

It is worth noting that the conditions for Theorem \ref{thm:1} and \ref{thm:2} depend only on the initial model gradient $\vg_2(\vw^1)$ before learning task 2 and the gradient $\vg_1(\vw^1)$ when learning task 1, which are both easily accessible and calculated. Particularly, the positive correlation can be evaluated by only storing the gradient $\vg_1(\vw^1)$ instead of the data of task 1. In stark contrast, the task gradient correlation characterization in \cite{chaudhry2018efficient, riemer2018learning} involves the gradient evaluation of old tasks with respect to the current model weight, and hence requires the data of old tasks when learning the new task.

\section{Continual learning with backward knowledge transfer}
\label{others}

Based on the theoretical analysis above, we next propose a continual learning method with backward knowledge transfer (CUBER), by selectively updating the learnt model of old tasks when learning the new task. In particular, CUBER works in a layer-wise manner: Given a $L$-layer network, CUBER first characterizes the task correlation for each layer, and then employs different strategies to learn the new task depending on the task correlation. 

More specifically, denote  the set of weights as $\vw=\{\vw_l\}_{l=1}^L$, where $\vw_l$ is the layer-wise weight for layer $l$.  Given the data input $\vx^t_{i}$ for task $t$, let $\vx^t_{l,i}$ be the input of layer $l$ and $\vx^t_{1,i}=\vx^t_{i}$.
Denote $f$ as the operation of the network layer.
The  output $\vx^t_{l+1,i}$ for  layer $l$ can be then computed as
    $\vx^t_{l+1,i}=f(\vw_l,\vx^t_{l,i})$.
Here we denote $\vx^t_{l,i}$ as the representations of the input $\vx^t_{i}$ at layer $l$. Given a new task $t\geq 2$, we characterize its task correlation with some old task $j\in[1, t-1]$ for layer $l$ into three different regimes based on Definition \ref{def:1} and \ref{def:2}. 

\begin{figure*}
\vspace{-0.1cm}
    \centering
    \includegraphics[width=0.8\linewidth]{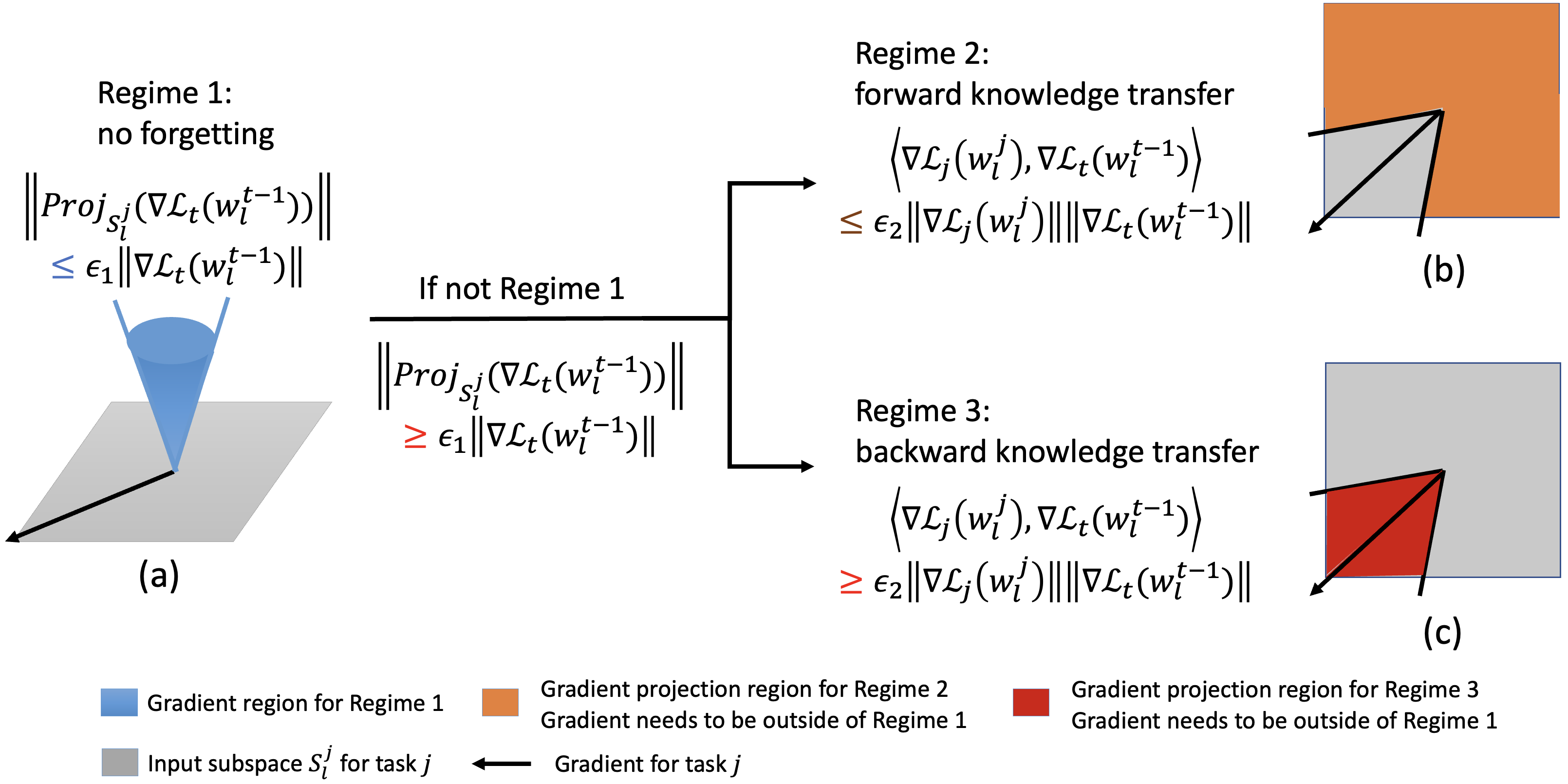}

    \caption{A simple illustration of the layer-wise task correlation detection. Given the new task $t$, an old task $j$ belongs to (1) Regime 1 if the initial model gradient $\nabla \gL_t(\vw_l^{t-1})$ of task $t$ has small projection onto the subspace $S_l^j$ of task $j$; (2) Regime 2 if the strong projection condition is satisfied while the projection of $\nabla \gL_t(\vw_l^{t-1})$ onto $S_l^j$ is not aligned well with the gradient of task $j$; (3) Regime 3 if both the strong projection condition and the positive correlation condition are satisfied.}
\label{fig:regime}
\vspace{-0.1cm}
\end{figure*}

\emph{Regime 1 (no forgetting):} We say that task $j\in Reg_{l,1}^t$ for layer $l$ if the following holds:
{\small
\begin{align*}
    \|\proj_{S^j_l}(\nabla \gL_t(\vw_l^{t-1}))\|\leq \epsilon_1\|\nabla \gL_t(\vw_l^{t-1}))\|.
\end{align*}}%
In this case, the layer-wise input subspace $S_l^j$ for task $j$ and $S_l^t$ for task $t$ are treated as nearly orthogonal ((a) in \cref{fig:regime}). As a result, there is little knowledge transfer between these two tasks, and updating the model along with $\nabla \gL_t(\vw_l)$ would not introduce much interference to task $j$. To reinforce the knowledge protection for task $j$, the model will be updated based on orthogonal projection:
{\small
\begin{align*}
    \nabla \gL_t(\vw_l) \longleftarrow \nabla \gL_t(\vw_l)-\proj_{S_l^j}(\nabla \gL_t(\vw_l)).
\end{align*}}%

\emph{Regime 2 (forward knowledge transfer)}: We say that task  $j\in Reg_{l,2}^t$ for layer $l$ if the following holds:
{\small
\begin{align*}
    \|\proj_{S^j_l}(\nabla \gL_t(\vw_l^{t-1}))\|&\geq \epsilon_1\|\nabla \gL_t(\vw_l^{t-1}))\|,\\
    \langle \nabla\gL_j(\vw_l^j), \nabla \gL_t(\vw_l^{t-1})\rangle&\leq \epsilon_2\|\nabla\gL_j(\vw_l^j)\|\|\nabla \gL_t(\vw_l^{t-1})\|.
\end{align*}}%
In this case, task $j$ and task $t$ can be strongly correlated for layer $l$ but possibly with `negative' correlation ((b) in \cref{fig:regime}), in the sense that updating the model along with $\nabla \gL_t(\vw_l)$ would substantially modify the learnt model $\proj_{S_l^j}(\vw^{t-1})$ for task $j$ in an unfavorable way and lead to the forgetting of task $j$. As there will be better forward knowledge transfer from the old task $j$ to the new task $t$ for layer $l$, we leverage the scaled weight projection in \cite{lin2022trgp} to facilitate the forward knowledge transfer through a scaling matrix $Q_l^{j,t}$, whereas the model is still updated using orthogonal projection to protect the knowledge of old tasks:
{\small
\begin{align}
\label{eq:scale}
     \nabla \gL_t(\vw_l) &\longleftarrow \nabla \gL_t(\vw_l)-\proj_{S_l^j}(\nabla \gL_t(\vw_l)),\nonumber\\
     \vspace{-0.2cm}
    Q_l^{j,t}&\longleftarrow Q_l^{j,t}- \beta \nabla_{Q} \gL_t(\vw_l-\proj_{S^j_{l}}(\vw_l)+\vw_l B^j_l Q_l^{j,t} (B^j_l)'). 
\end{align}}%
Here $B^j_l$ is the bases matrix for subspace $S_l^j$. Intuitively, the scaled weight projection $\vw_l B^j_l Q_l^{j,t} (B^j_l)'$ replaces the weight projection $\proj_{S^j_{l}}(\vw_l)$ of task $j$ by a scaled version, which transforms the knowledge of task $j$ to the appropriate model of the new task $t$ through the optimization of $Q_l^{j,t}$.

\emph{Regime 3 (backward knowledge transfer)}: We say task  $j\in Reg_{l,3}^t$ for layer $l$ if the following holds:
{\small
\begin{align*}
    \|\proj_{S^j_l}(\nabla \gL_t(\vw_l^{t-1}))\|&\geq \epsilon_1\|\nabla \gL_t(\vw_l^{t-1}))\|,\\
    \langle \nabla\gL_j(\vw_l^j), \nabla \gL_t(\vw_l^{t-1})\rangle&\geq \epsilon_2\|\nabla\gL_j(\vw_l^j)\|\|\nabla \gL_t(\vw_l^{t-1})\|.
\end{align*}
}%
With the sufficient projection and the positive correlation, updating the model along with $\nabla \gL_t(\vw_l)$ could possibly lead to a better model for continual learning and also improve the learning  performance of the old task $j$ ((c) in \cref{fig:regime}). To avoid overly-optimistic modification on the learnt model of task $j$, we further regularize the projection of the model change on the subspace $S_l^j$, given that the model projection is indeed frozen for task $j$ to address forgetting with orthogonal projection, i.e., $\proj_{S^j_{l}}(\vw_l^{t-1})=\proj_{S^j_{l}}(\vw_l^j)$.
This gives the following model update:
\begin{align*}
    \vw_l \longleftarrow \vw_l -\alpha \nabla [\gL_t(\vw_l)+\lambda\|\proj_{S^j_{l}}(\vw_l-\vw_l^{t-1})\|].
\end{align*}
Note that the gradient projection on the subspaces of old tasks in Regime 1 and 2 is removed from the gradient in the model update above.
Besides, we also learn a scaling matrix $Q_l^{j,t}$ for better forward knowledge transfer from the old task $j\in Reg_{l,3}^t$ to the new task $t$ as in \cref{eq:scale}.

\begin{wrapfigure}{r}{.32\textwidth}
\vspace{-0.1cm}
\includegraphics[width=0.2\textwidth]{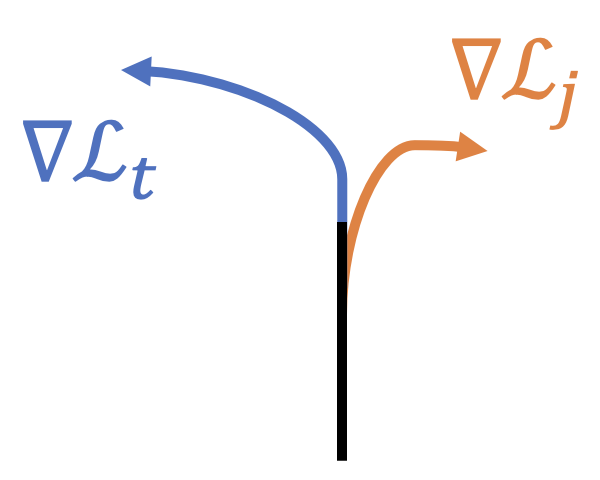}
\caption{ A simple illustration of the case where updating the model with $\nabla \gL_t$ benefits the old task $j\in Reg_{l,3}^t$ at the beginning but eventually conflicts with $\nabla \gL_j$.} 
\label{fig:reg3}
\vspace{-0.1cm}
\end{wrapfigure}
However, continuous model update with task $t$ gradient $\nabla \gL_t(\vw_l)$ will eventually lead to the task specific model for task $t$, which usually differs from the model of task $j$ in Regime 3 (\cref{fig:reg3}). To address this problem, note that the second part of Theorem \ref{thm:2} characterizes the condition under which updating the model with $\nabla \gL_t(\vw_l)$ will result in backward knowledge transfer. Thus motivated, for a model update iterate $\vw_{l,k}$ at $k$-th iteration when learning the new task $t$, we evaluate the following condition for task $j\in Reg_{l,3}^t$:
\begin{align}\label{eq:condition}
    \langle \nabla\gL_j(\vw_l^j), \nabla\gL_t(\vw_{l,k})\rangle \geq \epsilon_2\|\nabla\gL_j(\vw_l^j)\|\|\nabla \gL_t(\vw_{l,k})\|
\end{align}
and degenerate task $j$ to Regime 2, i.e., remove the gradient projection on the subspace $S_l^j$ from the task $t$ gradient and stop modifying the model for task $j$, if the condition (\ref{eq:condition}) does not hold. 

\emph{Bases extraction:} After learning the model $\vw^t$ for task $t$, we construct the input subspace for each layer $l$ by extracting bases from its representation $\vx^t_{l,i}$ based on singular value decomposition (SVD) \cite{lin2022trgp}. More specifically, given a batch of $n$ samples and the learnt model $\vw^t$, the representation matrix for layer $l$ is denoted as $R_l^t=[\vx^t_{l,1}, \vx^t_{l,2},...,\vx^t_{l,n}]$. Since the bases of the old tasks may include important bases for the new task $t$, we determine the bases for task $t$ from the union of the bases of the old tasks and the newly generated bases. Towards this end, we first concatenate the bases $B_l^j$ for $j\in [0, t-1]$ together in a matrix $O_l^t$ and remove the common bases. Then SVD is applied on the residual representation matrix $\Tilde{R}_l^t=R_l^t-R_l^t O_l^t (O_l^t)'$, i.e.,  $\Tilde{R}_l^t=U_l^t \Sigma_l^t (V_l^t)'$, where $U^t_l$ is the left singular matrix. We construct $B_l^t$ by selecting the most important bases from the pool of bases in $O^t_l$ and $U^t_l$ depending on their eigenvalues, which yields a low rank matrix approximation of $R_l^t$.

To conclude, the optimization problem for learning the new task $t$ can be summarized as follows:
{\small
\begin{align}
\label{eq:optim}
    \min_{\vw, \{Q_l^{j,t}\}_{l,j\in Reg^t_{l,2}\bigcup Reg^t_{l,3}}}~~~~&\gL_t(\{\Tilde{\vw}_l\}_l)+\lambda\sum_{l}\sum_{j\in Reg^t_{l,3}}  \|\proj_{S^j_{l}}(\vw_l-\vw_l^{t-1})\|,\\
    s.t.~~~~~~~~~~&\Tilde{\vw}_l=\vw_l+\sum_{j\in Reg^t_{l,2}\bigcup Reg^t_{l,3}} [\vw_l B^j_l Q_l^{j,t} (B^j_l)'-\proj_{S^j_{l}}(\vw_l)],\nonumber\\
    &\nabla \gL_t(\vw_l) = \nabla \gL_t(\vw_l)-\sum_{j\in Reg^t_{l,1}\bigcup Reg^t_{l,2}} \proj_{S_l^j}(\nabla \gL_t(\vw_l)).\nonumber
\end{align}}

The key idea is that we conservatively update the model for old tasks in Regime 3 while using orthogonal projection to preserve the knowledge of other old tasks; in the meanwhile, we leverage the scaled weight projection to reuse the model knowledge of old tasks in both Regime 2 and 3 to facilitate forward knowledge transfer. 
It is worth to note that the task correlation is determined before learning the new task $t$, as both the strong projection condition and the positive correlation condition only depend on the initial model gradient for the new task. And this can be achieved by a simple forward-backward pass through the initial model with a batch of new task data.
The overview of CUBER can be found in \cref{alg1}.

\begin{algorithm}[tb]
\footnotesize
	\caption{Continual learning with backward knowledge transfer (CUBER)}
	\label{alg1}
 	\begin{algorithmic}[1]
 	\State Input: task sequence $\sT=\{t\}_{t=1}^T$;
 	\State Learn the first task using vanilla stochastic gradient descent;
 	\State Extract the bases $\{B_l^1\}$ based on SVD using the learnt model $\vw^1$;
 	\For{each task $t$}
 	    \State Calculate gradient $\nabla \gL_t(\vw_{t-1})$;
 	    \State Evaluate the strong projection and the positive correlation conditions for layer-wise task correlation detection to determine $Reg_{l,1}^t$, $Reg_{l,2}^t$ and $Reg_{l,3}^t$;
 	    \For{k=1, 2,...}
 	        \State Update the model and scaling matrices by solving \cref{eq:optim};
 	        \For{task $j<t$ and $j\in Reg_{l,3}^t$}	            \If{$\langle \nabla\gL_j(\vw_l^j), \nabla\gL_t(\vw_{l,k})\rangle<\epsilon_2\|\nabla\gL_j(\vw_l^j)\|\|\nabla \gL_t(\vw_{l,k})\|$}
 	               \State Degenerate task j to $Reg_{l,2}^t$;
 	            \EndIf
 	        \EndFor
 	    \EndFor
 	    \State Store the gradient $\nabla \gL_t(\vw^t)$ in the task memory;
 	    \State Extract the bases $\{B_l^t\}$ based on SVD using the learnt model $\vw^t$; 
		\EndFor
	\end{algorithmic}

\end{algorithm}

\section{Experiments}

\textbf{Datasets.}
We evaluate the performance of CUBER on four standard CL benchmarks. (1) Permuted MNIST: a variant of the MNIST dataset \cite{lecun1998mnist} where random permutations are applied to the input pixels. Following \cite{lopez2017gradient, saha2021gradient}, we divide the dataset into 10 tasks with different permutations and each task includes 10 classes. (2) Split CIFAR-100: we divide the CIFAR-100 dataset \cite{krizhevsky2009learning} into 10 different tasks, where each task is a 10-way multi-class classification problem. (3) 5-Datasets: we consider a sequence of 5 datasets, i.e., CIFAR-10, MNIST, SVHN \cite{netzer2011reading}, not-MNIST\cite{bulatov2011notmnist}, Fashion MNIST\cite{xiao2017fashion}, and the classification problem on each dataset is a task. (4) Split MiniImageNet: we divide the MiniImageNet dataset \cite{vinyals2016matching} into 20 tasks, where each task includes 5 classes.

\textbf{Baselines.}
In this work, we compare CUBER with the following baselines on the benchmarks mentioned above. (1) EWC \cite{kirkpatrick2017overcoming}: a regularization-based method that leverages Fisher Information matrix for weights importance evaluation;  (2) HAT \cite{serra2018overcoming}: learns a hard attention mask to preserve the knowledge of old task; (3) Orthogonal Weight Modulation  (OWM) \cite{zeng2019continual}: learns a projector matrix to project the gradient of the new task to the orthogonal direction of input subspace of old tasks;  (4) Gradient Projection Memory (GPM) \cite{saha2021gradient}: stores the bases of the input subspace of old tasks and then updates the model with the gradient projection orthogonal to the subspace spanned by these bases; (5) TRGP \cite{lin2022trgp}: proposes a scaled weight projection to facilitate the forward knowledge transfer from related old tasks to the new task while updating the model based on orthogonal gradient projection, which demonstrates the state-of-the-art performance for a fixed capacity network; (6) Averaged GEM (A-GEM) \cite{chaudhry2018efficient}: constrains the new task learning with the gradient calculated using the stored data of old tasks; (7) Experience Replay with Reservior sample (ER\_Res) \cite{chaudhry2019continual}: uses a small episodic memory to store old task samples for addressing forgetting; (8) Multitask: jointly learns all tasks once with a single network using all datasets, which usually serves as a performance upper bound in CL \cite{saha2021gradient}.

\textbf{Network and training details.}
For a given dataset, we study all CL methods using the same network architecture. More specifically, for Permuted MNIST, we consider a 3-layer fully-connected network including 2 hidden layers with 100 units. And we train the network for 5 epochs with a batch size of 10 for every task. For Split CIFAR-100, we use a version of 5-layer AlexNet by following \cite{saha2021gradient, lin2022trgp}. When learning each task, we train the network for a maximum of 200 epochs with early termination based on the validation loss, and use a batch size of 64. For 5-Datasets, we use a reduced ResNet-18 \cite{lopez2017gradient} and follow the same training strategy as in Split CIFAR-100. For Split MiniImageNet, a reduced ResNet-18 is also used, and we train the network for a maximum of 100 epoches with early termination. The batch size is 64. Similar to \cite{lin2022trgp}, we select at most two tasks to be in Regime 2 and 3 for each layer with the largest gradient projection norm, to reduce the performance sensitivity on the choice of $\epsilon_1$. In the experiments, we set $\epsilon_1=0.5$.

To evaluate the learning performance, we consider the following two metrics, i.e., accuracy (ACC) which measures the final accuracy averaged over all tasks,  and backward transfer (BWT) which measures the average accuracy change of each task after learning new tasks:
{\small
\vspace{-0.1cm}
\begin{align*}
    ACC=\frac{1}{T}\sum\nolimits_{i=1}^T A_{T,i}, ~~ BWT=\frac{1}{T-1}\sum\nolimits_{i=1}^{T-1} (A_{T,i}-A_{i,i})
    \vspace{-0.1cm}
\end{align*}}%
where $A_{i,j}$ represents the testing accuracy of task $j$ after learning task $i$.

\subsection{Main results}

As shown in Table \ref{tab:deviation}, CUBER demonstrates the best performance of BWT on all datasets compared to the baselines. In particular, \emph{positive BWT can be obtained by CUBER on Split CIFAR-100 and Split MiniImageNet, which has not been achieved by previous works in a fixed capacity network without data-replay}. For 5-Dataset, since the tasks therein are less related to each other as in GPM \cite{saha2021gradient}, we do not expect much knowledge transfer across tasks (but knowledge transfer still exists in terms of the layer-wise features), and there is no forgetting in CUBER. Achieving zero-forgetting is very difficult 
for Permuted MNIST, because all the tasks share one output layer and there is no task identifier during testing (domain-incremental) \cite{saha2021gradient}. However, even in this case CUBER still achieves the best BWT, i.e, nearly non-forgetting, among all methods.
Clearly, the strong performance on BWT indicates that CUBER can effectively facilitate the backward knowledge transfer by wisely modifying the learnt model of old tasks.

Benefiting from the superior performance on the backward knowledge transfer, CUBER also achieves the best or comparable performance on the averaged accuracy. More specifically, CUBER improves around $1\%$ in ACC over the best prior results on Split CIFAR-100, Split MiniImageNet and Permuted MNIST, while showing the comparable performance with the state-of-the-art method TRGP even on 5-datasets where tasks are less correlated. Moreover, CUBER performs better than Multitask
on both 5-Dataset and Permuted MNIST, \emph{which implies the importance of studying knowledge transfer in CL:} Both TRGP and CUBER outperform Multitask on 5-Dataset because of the scaled weight projection to facilitate forward knowledge transfer, whereas by facilitating backward knowledge CUBER becomes the only method that outperforms Multitask on Permuted MNIST.


\begin{table*}[!htbp]
\scriptsize
\centering
\vspace{-0.1cm}
\caption{The ACC and BWT with the standard deviation values over 5 different runs on different datasets. Here for Split CIFAR-100, Split MiniImageNet and 5-Dataset we use a multi-head network, while a single-head network is used for Permuted MNIST. Moreover, $\epsilon_2=0.0$.}
\vspace{-0.1cm}
\label{tab:deviation}
\scalebox{0.81}{
\begin{tabular}{c|cccccc|cc}
\toprule
\multicolumn{1}{c|}{\multirow{3}{*}{Method}} & \multicolumn{6}{c|}{Multi-head} & \multicolumn{2}{c}{Domain-incremental}\\\cmidrule{2-9}
& \multicolumn{2}{c}{Split CIFAR-100} &   \multicolumn{2}{c}{Split MiniImageNet} &   \multicolumn{2}{c|}{5-Dataset} &   \multicolumn{2}{c}{Permuted MNIST}\\ \cmidrule{2-9}
 & ACC(\%) & BWT(\%)   &   ACC(\%) & BWT(\%)   & ACC(\%) & BWT(\%)  & ACC(\%) & BWT(\%)   \\ \midrule
Multitask            & $79.58\pm 0.54$   & -     &   $69.46\pm 0.62$    & -     &   $91.54\pm 0.28$   & -&    $96.70\pm 0.02$  & -     \\ \midrule
OWM                  & $50.94\pm 0.60$   & $-30\pm 1$   & -   & -  & -       & -    &$90.71\pm 0.11$   & $-1\pm 0$   \\
EWC                  & $68.80\pm 0.88$   & $-2\pm 1$   & $52.01\pm 2.53$   & $-12\pm 3$ &   $88.64\pm 0.26$   & $-4\pm 1$ &    $89.97\pm 0.57$   & $-4\pm 1$ \\
HAT                  & $72.06\pm 0.50$   & $0\pm 0$     & $59.78\pm 0.57$   & $-3\pm 0$  & $91.32 \pm 0.18$  & $-1\pm 0$  & -   & -\\
A-GEM                & $63.98\pm 1.22$   & $-15\pm 2$   & $57.24\pm 0.72$   & $-12\pm 1$   & $84.04\pm 0.33$   & $-12\pm 1$   &  $83.56 \pm 0.16$   & $-14\pm 1$\\
ER\_Res              & $71.73\pm 0.63$   & $-6\pm 1$   & $58.94\pm 0.85$   & $-7\pm 1$   & $88.31\pm 0.22$   & $-4\pm 0$   &  $87.24\pm  0.53$  & $-11\pm 1$\\
GPM                  & $72.48\pm 0.40$   & $-0.9 \pm 0$  & $60.41\pm 0.61$   & $-0.7\pm 0.4$   & $91.22\pm 0.20$   &   $-1\pm 0 $   &  $93.91\pm 0.16$   & $-3\pm 0$ \\
TRGP & $74.46\pm 0.32$ & $-0.9\pm 0.01$  &$61.78\pm 0.60$ &  $-0.5\pm0.6 $  & $\pmb{93.56}\pm 0.10$ & $-0.04\pm 0.01$  &  $96.34\pm 0.11$ & $-0.8\pm 0.1$
\\ \midrule
CUBER (ours)                 &
$\pmb{75.54} \pm 0.22$ & $\pmb{0.13} \pm 0.08$  &    
$\pmb{62.67} \pm 0.74$ & $\pmb{0.23} \pm 0.15$  &    
$\pmb{93.48} \pm 0.10 $& $\pmb{0.00} \pm 0.02 $ &      
  $\pmb{97.25} \pm 0.00$ & $\pmb{-0.02} \pm 0.00$  \\ \bottomrule
\end{tabular}
}
\end{table*}

\subsection{Ablation studies}
\vspace{-0.05cm}
\begin{table}[!htbp]
\scriptsize
\vspace{-0.1cm}
\caption{The comparison between CUBER and TRGP in OL-CIFAR100. The selected old task (*) in (b) represents the old task with the largest number of layers in Regime 3 of the new task.}
\label{tab:back}
\vspace{-0.1cm}
\begin{subtable}{.35\linewidth}
\caption{The comparison of ACC and BWT.}
\label{tab:acc}
\centering
\scalebox{0.85}{
\begin{tabular}{c|cccc}
\toprule
\multicolumn{1}{c|}{\multirow{1}{*}{Method}} & \multicolumn{1}{c}{\multirow{1}{*}{ACC($\%$)}}&
\multicolumn{1}{c}{\multirow{1}{*}{BWT($\%$)}}\\
\midrule
     CUBER& 74.94 & 0.28\\\midrule
     TRGP& 74.33 & -0.18
     \\ \bottomrule
\end{tabular}}
\end{subtable}%
\begin{subtable}{.65\linewidth}
\caption{BWT-S of the selected old tasks.}
\label{tab:select}
\centering
\scalebox{0.85}{
\begin{tabular}{c|cccccccc}
\toprule
& Task 1 & Task 2 & Task 4  &  Task 5 & Average
\\ \midrule
selected old task* & Task 0 & Task 1 & Task 3 & Task 0 & -  
\\\midrule 
BWT-S (CUBER)     & 0.50      &   0.71      & 0.03       &   0.07      & 0.33      \\\midrule    
BWT-S (TRGP)     & -0.20      &   0.10      & -0.60       &   -0.20      & -0.23   
\\ \bottomrule
\end{tabular}
}
\end{subtable}

\end{table}

\textbf{Backward knowledge transfer.} 
The value of backward knowledge transfer indicates the average accuracy improvement for each old task after learning all tasks, which implies that the new task learning provides additional useful information for learning features of similar old tasks. Intuitively, this value should depend on the task similarity in CL.
To better understand the advantage of CUBER in terms of backward knowledge transfer, we further consider a special setup which includes a sequence of similar and dissimilar tasks. Specifically, different with Split-CIFAR100 where no tasks have overlapping classes, we split the first 50 classes in CIFAR100 into 7 tasks (OL-CIFAR100): Task 0-6 contain classes 0-9, 5-14, 10-19, 20-29, 25-34, 30-39, 40-49, respectively. We compare the performance of CUBER with TRGP in Table \ref{tab:back}. As shown in Table \ref{tab:acc}, CUBER clearly outperforms TRGP in both ACC and BWT. In particular, CUBER has a positive BWT of $0.28\%$, where TRGP suffers from forgetting. We further analyze the task selections in Table \ref{tab:select}, where ``selected old task'' refers to the task that has the largest number of layers in Regime 3 of the new task. For example, according to the setup of OL-CIFAR100, the selected old task for Task 4 should be Task 3 with a high probability as they have overlapping classes. And we denote a new metric, namely BWT-S, to evaluate the backward knowledge transfer of the selected old task after learning the new task, i.e., BWT-S=$ A_{t,j}-A_{t-1,j}$ where $j$ is the ``selected old task'' of the new task $t$. As shown in Table \ref{tab:select}, CUBER correctly identifies the correlated tasks for most new tasks except Task 5. This is a reasonable result because the task correlation detection is based on the initial model gradient which is noisy in general, and hence only serves as an estimation of underlying true correlation. However, such an estimated task correlation characterization is indeed sufficient to effectively facilitate backward knowledge transfer, as corroborated by the positive BWT-S and the superior performance of CUBER.

\textbf{Forward knowledge transfer.} We also evaluate the forward knowledge transfer (FWT) in CUBER, compared to the best two baseline methods GPM and TRGP. Here the FWT measures the gap between $A_{i,i}$ and the accuracy of learning task $i$ only from scratch. For simplicity, we use the FWT of GPM as a baseline and evaluate the improvements of TRGP and CUBER over GPM. It can be seen from Table \ref{tab:forward} that CUBER performs even better than TRGP although CUBER follows the same strategy, i.e., the scaled weight projection, to facilitate the forward knowledge transfer,
and achieves the best FWT among the three methods in most cases. The reason behind is that the characterization of Regime 3 in CUBER not only allows the modification of the learnt model of the old tasks to prompt the backward knowledge transfer, but also relaxes the  constraint on the gradient update for the new task, i.e., the model can be now updated in the subspace of the selected old tasks for learning the new task. This gradient constraint relaxation consequently leads to better model learning of the new task.

\vspace{-0.05cm}
\begin{table}[!htbp]
\scriptsize
\vspace{-0.1cm}
\caption{Comparison of FWT among GPM, TRGP and CUBER. The value for GPM is zero because we treat GPM as the baseline and consider the relative FWT improvement over GPM.}
\label{tab:forward}
\centering
\scalebox{0.85}{
\begin{tabular}{c|cccc}
\toprule
Method & Split CIFAR-100 & Split MiniImageNet & 5-Dataset & Permuted MNIST\\
\midrule
     GPM & 0 & 0 & 0 & 0\\\midrule
     TRGP & 2.01 & 2.36 &1.98 & 0.18 \\\midrule
     CUBER & 2.79 & 3.13 & 1.96 &0.8
     \\ \bottomrule
\end{tabular}}
\vspace{-0.1cm}
\end{table}

\textbf{Impact of $\epsilon_2$.} It is clear that the selection of layer-wise Regime 3 depends on the value of the threshold $\epsilon_2$. To show the impact of $\epsilon_2$, we evaluate the learning performance under different values of $\epsilon_2$ in Split CIFAR-100. As shown in Table \ref{tab:ep}, the performance on ACC is comparable for all three cases and the BWT decreases as the value of $\epsilon_2$ increases. Intuitively, $\epsilon_2$ characterizes the conservatism in selecting tasks to Regime 3 and modifying the model of selected tasks. Specifically, with a larger $\epsilon_2$, we just consider the backward knowledge transfer to the old tasks that are strongly correlated with the new task, and only slightly modify the learnt model of these selected tasks, because the condition \cref{eq:condition} can be quickly violated with the model update and CUBER will stop changing the learnt model of the selected tasks. 

\vspace{-0.05cm}
\begin{table}[!htbp]
\scriptsize
\vspace{-0.1cm}
\caption{The impact of $\epsilon_2$ on the performance in Split CIFAR-100.}
\label{tab:ep}
\centering
\scalebox{0.85}{
\begin{tabular}{ccccccccc}
\toprule
\multicolumn{2}{c}{$\epsilon_2=0.0$} &  & \multicolumn{2}{c}{$\epsilon_2=0.2$} &  & \multicolumn{2}{c}{$\epsilon_2=0.5$} 
\\ \cmidrule{1-2} \cmidrule{4-5} \cmidrule{7-8} 
ACC(\%)   & BWT(\%)   &  & ACC(\%)    & BWT(\%)    &  & ACC(\%)    & BWT(\%)    
\\\midrule 
75.54     & 0.22      &  & 75.73      & 0.03       &  & 75.55      & 0.01       
\\ \bottomrule
\end{tabular}
}
\vspace{-0.1cm}
\end{table}

\section{Conclusion}

In this work, we study the problem of backward knowledge transfer in CL. Different from most existing methods that generally freeze the learnt model of the old tasks so as to mitigate catastrophic forgetting, this study seeks to carefully modify the learnt model to facilitate backward knowledge transfer from the new task to the old tasks. To this end, we first introduce notions of strong projection and positive correlation to characterize the task correlation, and show that when the task gradients are sufficiently aligned in the old task subspace, appropriate model change for the old tasks could be beneficial for CL and result in better backward knowledge transfer. Based on the theoretical analysis, we next propose CUBER to carefully learn the model for the new task, which would carefully 
update the learnt model of the old tasks that are positively correlated with the new task. As shown in the experimental results, CUBER can successfully improve the backward knowledge transfer on the standard CL benchmarks in contrast to related baselines. 

\textbf{Impact and limitations.~} The mainstream strategy nowadays to address forgetting is to minimize the interference to old tasks and avoid the learnt model change, which may however conflict with the goal of CL, in the sense that the backward knowledge transfer from the new task to old tasks can be restricted without modifying the learnt model. In this work, we go beyond this strategy and shed light on the relationship between backward knowledge transfer and model modification, by characterizing the task correlations with gradient projection. We hope that this work will serve as initial steps and motivate further research in CL community on the important while less explored problem, i.e., how to design algorithms that can provide targeted treatments to achieve backward knowledge transfer.
However, CUBER also comes with several limitations.
As in recent orthogonal-projection based CL methods,
CUBER extracts the bases of the task subspaces based on SVD, which may lead to high computational cost for large dimensional data. How to reduce the complexity is an interesting direction. Another limitation is that we assume that clear task boundaries exist between different tasks. In future work, it is of great interests to extend CUBER to more general CL settings.

\section*{Acknowledgement}
The work of S. Lin and J. Zhang was supported in part by the U.S. National Science Foundation Grants CNS-2203239,  CNS-2203412,  RINGS-2148253,  and CCSS-2203238. The work of L. Yang 
and D. Fan was supported in part by the U.S. National Science Foundation Grants No. 1931871 and No. 2144751.

\bibliography{reference}

\begin{thebibliography}{10}

\bibitem{aljundi2018memory}
Rahaf Aljundi, Francesca Babiloni, Mohamed Elhoseiny, Marcus Rohrbach, and
  Tinne Tuytelaars.
\newblock Memory aware synapses: Learning what (not) to forget.
\newblock In {\em Proceedings of the European Conference on Computer Vision
  (ECCV)}, pages 139--154, 2018.

\bibitem{bulatov2011notmnist}
Yaroslav Bulatov.
\newblock Notmnist dataset.
\newblock {\em Google (Books/OCR), Tech. Rep.[Online]. Available:
  http://yaroslavvb. blogspot. it/2011/09/notmnist-dataset. html}, 2, 2011.

\bibitem{chaudhry2018efficient}
Arslan Chaudhry, Marc'Aurelio Ranzato, Marcus Rohrbach, and Mohamed Elhoseiny.
\newblock Efficient lifelong learning with a-gem.
\newblock {\em arXiv preprint arXiv:1812.00420}, 2018.

\bibitem{chaudhry2019continual}
Arslan Chaudhry, Marcus Rohrbach, Mohamed Elhoseiny, Thalaiyasingam Ajanthan,
  Puneet~K Dokania, Philip~HS Torr, and M~Ranzato.
\newblock Continual learning with tiny episodic memories.
\newblock 2019.

\bibitem{chen2018lifelong}
Zhiyuan Chen and Bing Liu.
\newblock Lifelong machine learning.
\newblock {\em Synthesis Lectures on Artificial Intelligence and Machine
  Learning}, 12(3):1--207, 2018.

\bibitem{farajtabar2020orthogonal}
Mehrdad Farajtabar, Navid Azizan, Alex Mott, and Ang Li.
\newblock Orthogonal gradient descent for continual learning.
\newblock In {\em International Conference on Artificial Intelligence and
  Statistics}, pages 3762--3773. PMLR, 2020.

\bibitem{fernando2017pathnet}
Chrisantha Fernando, Dylan Banarse, Charles Blundell, Yori Zwols, David Ha,
  Andrei~A Rusu, Alexander Pritzel, and Daan Wierstra.
\newblock Pathnet: Evolution channels gradient descent in super neural
  networks.
\newblock {\em arXiv preprint arXiv:1701.08734}, 2017.

\bibitem{guo2020improved}
Yunhui Guo, Mingrui Liu, Tianbao Yang, and Tajana Rosing.
\newblock Improved schemes for episodic memory based lifelong learning
  algorithm.
\newblock In {\em Conference on Neural Information Processing Systems}, 2020.

\bibitem{hung2019compacting}
Steven~CY Hung, Cheng-Hao Tu, Cheng-En Wu, Chien-Hung Chen, Yi-Ming Chan, and
  Chu-Song Chen.
\newblock Compacting, picking and growing for unforgetting continual learning.
\newblock {\em arXiv preprint arXiv:1910.06562}, 2019.

\bibitem{kao2021natural}
Ta-Chu Kao, Kristopher Jensen, Gido van~de Ven, Alberto Bernacchia, and
  Guillaume Hennequin.
\newblock Natural continual learning: success is a journey, not (just) a
  destination.
\newblock {\em Advances in Neural Information Processing Systems},
  34:28067--28079, 2021.

\bibitem{ke2020continual}
Zixuan Ke, Bing Liu, and Xingchang Huang.
\newblock Continual learning of a mixed sequence of similar and dissimilar
  tasks.
\newblock {\em Advances in Neural Information Processing Systems},
  33:18493--18504, 2020.

\bibitem{kirkpatrick2017overcoming}
James Kirkpatrick, Razvan Pascanu, Neil Rabinowitz, Joel Veness, Guillaume
  Desjardins, Andrei~A Rusu, Kieran Milan, John Quan, Tiago Ramalho, Agnieszka
  Grabska-Barwinska, et~al.
\newblock Overcoming catastrophic forgetting in neural networks.
\newblock {\em Proceedings of the national academy of sciences},
  114(13):3521--3526, 2017.

\bibitem{krizhevsky2009learning}
Alex Krizhevsky, Geoffrey Hinton, et~al.
\newblock Learning multiple layers of features from tiny images.
\newblock 2009.

\bibitem{lecun1998mnist}
Yann LeCun.
\newblock The mnist database of handwritten digits.
\newblock {\em http://yann. lecun. com/exdb/mnist/}, 1998.

\bibitem{lee2017overcoming}
Sang-Woo Lee, Jin-Hwa Kim, Jaehyun Jun, Jung-Woo Ha, and Byoung-Tak Zhang.
\newblock Overcoming catastrophic forgetting by incremental moment matching.
\newblock {\em arXiv preprint arXiv:1703.08475}, 2017.

\bibitem{li2017learning}
Zhizhong Li and Derek Hoiem.
\newblock Learning without forgetting.
\newblock {\em IEEE transactions on pattern analysis and machine intelligence},
  40(12):2935--2947, 2017.

\bibitem{lin2022trgp}
Sen Lin, Li~Yang, Deliang Fan, and Junshan Zhang.
\newblock Trgp: Trust region gradient projection for continual learning.
\newblock {\em Tenth International Conference on Learning Representations, ICLR
  2022}, 2022.

\bibitem{liu2022continual}
Hao Liu and Huaping Liu.
\newblock Continual learning with recursive gradient optimization.
\newblock {\em arXiv preprint arXiv:2201.12522}, 2022.

\bibitem{lopez2017gradient}
David Lopez-Paz and Marc'Aurelio Ranzato.
\newblock Gradient episodic memory for continual learning.
\newblock {\em Advances in neural information processing systems},
  30:6467--6476, 2017.

\bibitem{mccloskey1989catastrophic}
Michael McCloskey and Neal~J Cohen.
\newblock Catastrophic interference in connectionist networks: The sequential
  learning problem.
\newblock In {\em Psychology of learning and motivation}, volume~24, pages
  109--165. Elsevier, 1989.

\bibitem{netzer2011reading}
Yuval Netzer, Tao Wang, Adam Coates, Alessandro Bissacco, Bo~Wu, and Andrew~Y
  Ng.
\newblock Reading digits in natural images with unsupervised feature learning.
\newblock 2011.

\bibitem{riemer2018learning}
Matthew Riemer, Ignacio Cases, Robert Ajemian, Miao Liu, Irina Rish, Yuhai Tu,
  and Gerald Tesauro.
\newblock Learning to learn without forgetting by maximizing transfer and
  minimizing interference.
\newblock {\em arXiv preprint arXiv:1810.11910}, 2018.

\bibitem{ring1994continual}
Mark~Bishop Ring et~al.
\newblock Continual learning in reinforcement environments.
\newblock 1994.

\bibitem{rusu2016progressive}
Andrei~A Rusu, Neil~C Rabinowitz, Guillaume Desjardins, Hubert Soyer, James
  Kirkpatrick, Koray Kavukcuoglu, Razvan Pascanu, and Raia Hadsell.
\newblock Progressive neural networks.
\newblock {\em arXiv preprint arXiv:1606.04671}, 2016.

\bibitem{saha2021gradient}
Gobinda Saha, Isha Garg, and Kaushik Roy.
\newblock Gradient projection memory for continual learning.
\newblock In {\em International Conference on Learning Representations}, 2021.

\bibitem{serra2018overcoming}
Joan Serra, Didac Suris, Marius Miron, and Alexandros Karatzoglou.
\newblock Overcoming catastrophic forgetting with hard attention to the task.
\newblock In {\em International Conference on Machine Learning}, pages
  4548--4557. PMLR, 2018.

\bibitem{vinyals2016matching}
Oriol Vinyals, Charles Blundell, Timothy Lillicrap, Daan Wierstra, et~al.
\newblock Matching networks for one shot learning.
\newblock {\em Advances in neural information processing systems}, 29, 2016.

\bibitem{xiao2017fashion}
Han Xiao, Kashif Rasul, and Roland Vollgraf.
\newblock Fashion-mnist: a novel image dataset for benchmarking machine
  learning algorithms.
\newblock {\em arXiv preprint arXiv:1708.07747}, 2017.

\bibitem{yang2021grown}
Li~Yang, Sen Lin, Junshan Zhang, and Deliang Fan.
\newblock Grown: Grow only when necessary for continual learning.
\newblock {\em arXiv preprint arXiv:2110.00908}, 2021.

\bibitem{yoon2020scalable}
Jaehong Yoon, Saehoon Kim, Eunho Yang, and Sung~Ju Hwang.
\newblock Scalable and order-robust continual learning with additive parameter
  decomposition.
\newblock In {\em Eighth International Conference on Learning Representations,
  ICLR 2020}. ICLR, 2020.

\bibitem{yoon2017lifelong}
Jaehong Yoon, Eunho Yang, Jeongtae Lee, and Sung~Ju Hwang.
\newblock Lifelong learning with dynamically expandable networks.
\newblock {\em arXiv preprint arXiv:1708.01547}, 2017.

\bibitem{zeng2019continual}
Guanxiong Zeng, Yang Chen, Bo~Cui, and Shan Yu.
\newblock Continual learning of context-dependent processing in neural
  networks.
\newblock {\em Nature Machine Intelligence}, 1(8):364--372, 2019.

\bibitem{zhang2021understanding}
Chiyuan Zhang, Samy Bengio, Moritz Hardt, Benjamin Recht, and Oriol Vinyals.
\newblock Understanding deep learning (still) requires rethinking
  generalization.
\newblock {\em Communications of the ACM}, 64(3):107--115, 2021.

\end{thebibliography}
\bibliographystyle{plain}

\section*{Checklist}

The checklist follows the references.  Please
read the checklist guidelines carefully for information on how to answer these
questions.  For each question, change the default \answerTODO{} to \answerYes{},
\answerNo{}, or \answerNA{}.  You are strongly encouraged to include a {\bf
justification to your answer}, either by referencing the appropriate section of
your paper or providing a brief inline description.  For example:
\begin{itemize}
  \item Did you include the license to the code and datasets? \answerYes{See Section~\ref{gen_inst}.}
  \item Did you include the license to the code and datasets? \answerNo{The code and the data are proprietary.}
  \item Did you include the license to the code and datasets? \answerNA{}
\end{itemize}
Please do not modify the questions and only use the provided macros for your
answers.  Note that the Checklist section does not count towards the page
limit.  In your paper, please delete this instructions block and only keep the
Checklist section heading above along with the questions/answers below.

\begin{enumerate}

\item For all authors...
\begin{enumerate}
  \item Do the main claims made in the abstract and introduction accurately reflect the paper's contributions and scope?
    \answerYes{}
  \item Did you describe the limitations of your work?
    \answerYes{See Section 6.}
  \item Did you discuss any potential negative societal impacts of your work?
    \answerYes{See Section 6.}
  \item Have you read the ethics review guidelines and ensured that your paper conforms to them?
    \answerYes{}
\end{enumerate}

\item If you are including theoretical results...
\begin{enumerate}
  \item Did you state the full set of assumptions of all theoretical results?
    \answerYes{See Theorem 1 and 2.}
        \item Did you include complete proofs of all theoretical results?
    \answerYes{See the appendix.}
\end{enumerate}

\item If you ran experiments...
\begin{enumerate}
  \item Did you include the code, data, and instructions needed to reproduce the main experimental results (either in the supplemental material or as a URL)?
    \answerYes{We include the code in the supplemental material.}
  \item Did you specify all the training details (e.g., data splits, hyperparameters, how they were chosen)?
    \answerYes{See section 5.}
        \item Did you report error bars (e.g., with respect to the random seed after running experiments multiple times)?
    \answerYes{See section 5.}
        \item Did you include the total amount of compute and the type of resources used (e.g., type of GPUs, internal cluster, or cloud provider)?
    \answerYes{See the appendix.}
\end{enumerate}

\item If you are using existing assets (e.g., code, data, models) or curating/releasing new assets...
\begin{enumerate}
  \item If your work uses existing assets, did you cite the creators?
    \answerYes{See section 5.}
  \item Did you mention the license of the assets?
    \answerYes{See the appendix.}
  \item Did you include any new assets either in the supplemental material or as a URL?
    \answerYes{We include code in the supplemental material.}
  \item Did you discuss whether and how consent was obtained from people whose data you're using/curating?
    \answerNA{}
  \item Did you discuss whether the data you are using/curating contains personally identifiable information or offensive content?
    \answerNA{}
\end{enumerate}

\item If you used crowdsourcing or conducted research with human subjects...
\begin{enumerate}
  \item Did you include the full text of instructions given to participants and screenshots, if applicable?
    \answerNA{}
  \item Did you describe any potential participant risks, with links to Institutional Review Board (IRB) approvals, if applicable?
    \answerNA{}
  \item Did you include the estimated hourly wage paid to participants and the total amount spent on participant compensation?
    \answerNA{}
\end{enumerate}

\end{enumerate}


\newpage
\appendix

\section*{Appendix}

\section{Proof of Theorem \ref{thm:1}}


For a $H/2$-smooth loss function $\gL$, it can be easily shown that $\gF$ is $H$-smooth.

(1) For any $k\in[0, K]$, we can have
\begin{align}\label{smooth}
    \gF(\vw_{k+1})\leq & \gF(\vw_k)+\nabla \gF(\vw_k)^T (\vw_{k+1}-\vw_k) + \frac{H}{2}\|\vw_{k+1}-\vw_k\|^2\nonumber\\
    =& \gF(\vw_k)+ (\vg_1(\vw_k)+\vg_2(\vw_k))^T(-\alpha \vg_2(\vw_k))+\frac{\alpha^2 H}{2}\|\vg_2(\vw_k)\|^2\nonumber\\
    =& \gF(\vw_k)-\left[\alpha-\frac{\alpha^2 H}{2}\right]\|\vg_2(\vw_k)\|^2-\alpha\langle \vg_1(\vw_k),\vg_2(\vw_k) \rangle.
\end{align}

For the term $\langle \vg_1(\vw_k),\vg_2(\vw_k) \rangle$, it follows that
\begin{align}\label{inner}
    &\langle \vg_1(\vw_k),\vg_2(\vw_k) \rangle\nonumber\\
    =& \langle \vg_1(\vw_k)-\vg_1(\vw_0)+\vg_1(\vw_0),\vg_2(\vw_k) \rangle\nonumber\\
    =& \langle \vg_1(\vw_k)-\vg_1(\vw_0),\vg_2(\vw_k) \rangle +\langle \vg_1(\vw_0),\vg_2(\vw_k) \rangle\nonumber\\
    =& \langle \vg_1(\vw_k)-\vg_1(\vw_0),\vg_2(\vw_k) \rangle +\langle \vg_1(\vw_0),\vg_2(\vw_k)-\vg_2(\vw_0) \rangle +\langle \vg_1(\vw_0),\vg_2(\vw_0) \rangle.
\end{align}

Because
\begin{align*}
    &2\langle \vg_1(\vw_k)-\vg_1(\vw_0),\vg_2(\vw_k) \rangle+\|\vg_1(\vw_k)-\vg_1(\vw_0)\|^2+\|\vg_2(\vw_k)\|^2\\
    =&\|\vg_1(\vw_k)-\vg_1(\vw_0)+\vg_2(\vw_k)\|^2\geq 0,
\end{align*}
we have
\begin{align}\label{inner_1}
    \langle \vg_1(\vw_k)-\vg_1(\vw_0),\vg_2(\vw_k) \rangle\geq -\frac{1}{2}\|\vg_1(\vw_k)-\vg_1(\vw_0)\|^2-\frac{1}{2}\|\vg_2(\vw_k)\|^2.
\end{align}

Following the same line, it can be shown that 
\begin{align}\label{inner_2}
    \langle \vg_1(\vw_0),\vg_2(\vw_k)-\vg_2(\vw_0)\rangle\geq -\frac{1}{2}\|\vg_2(\vw_k)-\vg_2(\vw_0)\|^2-\frac{1}{2}\|\vg_1(\vw_0)\|^2.
\end{align}

Combining \cref{inner}, \cref{inner_1} and \cref{inner_2} gives a lower bound on $\langle \vg_1(\vw_k),\vg_2(\vw_k) \rangle$, i.e.,
\begin{align}\label{eq:inner}
    &\langle \vg_1(\vw_k),\vg_2(\vw_k) \rangle\nonumber\\
    \geq & -\frac{1}{2}\|\vg_1(\vw_k)-\vg_1(\vw_0)\|^2-\frac{1}{2}\|\vg_2(\vw_k)\|^2\nonumber\\
    &-\frac{1}{2}\|\vg_2(\vw_k)-\vg_2(\vw_0)\|^2-\frac{1}{2}\|\vg_1(\vw_0)\|^2+\langle \vg_1(\vw_0),\vg_2(\vw_0) \rangle\nonumber\\
    \geq & -\frac{H^2}{8}\|\vw_k-\vw_0\|^2-\frac{1}{2}\|\vg_2(\vw_k)\|^2\nonumber\\
    &-\frac{H^2}{8}\|\vw_k-\vw_0\|^2-\frac{1}{2}\|\vg_1(\vw_0)\|^2+\langle \vg_1(\vw_0),\vg_2(\vw_0) \rangle\nonumber\\
    =&-\frac{H^2}{4}\|\vw_k-\vw_0\|^2-\frac{1}{2}\|\vg_2(\vw_k)\|^2-\frac{1}{2}\|\vg_1(\vw_0)\|^2+\langle \vg_1(\vw_0),\vg_2(\vw_0) \rangle,
\end{align}
where the second inequality is true because of the smoothness of the loss function.

Based on the update Rule $\#2$, it can be seen that
\begin{align}\label{eq:update}
    \vw_k = \vw_0 -\alpha\sum_{i=0}^{k-1} \vg_2(\vw_i).
\end{align}

Therefore, continuing with \cref{smooth}, we can have
\begingroup
\allowdisplaybreaks
\begin{align*}
    &\gF(\vw_{k+1})\\
    \leq &\gF(\vw_k)-\left[\alpha-\frac{\alpha^2 H}{2}\right]\|\vg_2(\vw_k)\|^2-\alpha\langle \vg_1(\vw_k),\vg_2(\vw_k) \rangle\\
    \leq &\gF(\vw_k)-\left[\alpha-\frac{\alpha^2 H}{2}\right]\|\vg_2(\vw_k)\|^2 + \frac{\alpha^3 H^2}{4} \|\sum_{i=0}^{k-1} \vg_2(\vw_i)\|^2+ \frac{\alpha}{2}\|\vg_2(\vw_k)\|^2\\
    &+\frac{\alpha}{2}\|\vg_1(\vw_0)\|^2-\alpha\langle \vg_1(\vw_0),\vg_2(\vw_0) \rangle\\
    =& \gF(\vw_k)-\left[\frac{\alpha}{2}-\frac{\alpha^2 H}{2}\right]\|\vg_2(\vw_k)\|^2+ \frac{\alpha^3 H^2}{4} \|\sum_{i=0}^{k-1} \vg_2(\vw_i)\|^2+\frac{\alpha}{2}\|\vg_1(\vw_0)\|^2-\alpha\langle \vg_1(\vw_0),\vg_2(\vw_0) \rangle\\
    \leq & \gF(\vw_k)-\left[\frac{\alpha}{2}-\frac{\alpha^2 H}{2}\right]\|\vg_2(\vw_k)\|^2+ \frac{\alpha^3 H^2}{4} \|\sum_{i=0}^{k-1} \vg_2(\vw_i)\|^2+\frac{\alpha}{2}\|\vg_1(\vw_0)\|^2\\
    &-\alpha\epsilon_2\|\vg_1(\vw_0)\|\|\vg_2(\vw_0)\|,
\end{align*}
\endgroup
where the last inequality is based on Definition \ref{def:2}.

Next, it can be shown that
\begin{align*}
    \alpha\leq \frac{\gamma \|\vg_1(\vw_0)\|}{HBK}\leq \frac{\gamma \|\vg_1(\vw_0)\|}{H\|\sum_{i=0}^{k-1} \vg_2(\vw_i)\|}.
\end{align*}

It then follows that
\begin{align}\label{sum}
    &\frac{1}{2}\|\vg_1(\vw_0)\|^2+\frac{\alpha^2 H^2}{4} \|\sum_{i=0}^{k-1} \vg_2(\vw_i)\|^2\nonumber\\
    \leq& \frac{1}{2}\|\vg_1(\vw_0)\|^2+ \frac{\gamma^2 \|\vg_1(\vw_0)\|^2}{4H^2\|\sum_{i=0}^{k-1} \vg_2(\vw_i)\|^2}H^2 \|\sum_{i=0}^{k-1} \vg_2(\vw_i)\|^2\nonumber\\
    =&\frac{2+\gamma^2}{4}\|\vg_1(\vw_0)\|^2.
\end{align}

Therefore, we can obtain that
\begin{align*}
    \gF(\vw_{k+1})\leq& \gF(\vw_k)-\left[\frac{\alpha}{2}-\frac{\alpha^2 H}{2}\right]\|\vg_2(\vw_k)\|^2+\frac{\alpha(2+\gamma^2)}{4}\|\vg_1(\vw_0)\|^2-\alpha\epsilon_2\|\vg_1(\vw_0)\|\|\vg_2(\vw_0)\|\\
    \leq& \gF(\vw_k)-\left[\frac{\alpha}{2}-\frac{\alpha^2 H}{2}\right]\|\vg_2(\vw_k)\|^2\\
    <&\gF(\vw_k),
\end{align*}
where the second inequality is true because $\epsilon_2\geq \frac{(2+\gamma^2)\|\vg_1(\vw_0)\|}{4\|\vg_2(\vw_0)\|}$.
This sufficient decrease of the objective function value indicates that the optimal $\gF(w^*)$ can be obtained eventually for convex loss functions.

(2) For a  non-convex loss function $\gL$, we can have the following as in \cref{smooth}:
\begingroup
\allowdisplaybreaks
\begin{align*}
    \gF(w_{k+1}^r)\leq& \gF(\vw_k)-\left[\alpha-\frac{\alpha^2 H}{2}\right]\|\vg_2(\vw_k)\|^2-\alpha\langle \vg_1(\vw_k),\vg_2(\vw_k) \rangle\\
    \overset{(a)}{=}& \gF(\vw_k)-\left[\alpha-\frac{\alpha^2 H}{2}\right]\|\vg_2(\vw_k)\|^2-\frac{\alpha}{2}[\|\nabla \gF(\vw_k)\|^2-\|\vg_1(\vw_k)\|^2-\|\vg_2(\vw_k)\|^2]\\
    =& \gF(\vw_k)-\left[\frac{\alpha}{2}-\frac{\alpha^2 H}{2}\right]\|\vg_2(\vw_k)\|^2-\frac{\alpha}{2}\|\nabla \gF(\vw_k)\|^2+\frac{\alpha}{2}\|\vg_1(\vw_k)\|^2\\
    =& \gF(\vw_k)-\left[\frac{\alpha}{2}-\frac{\alpha^2 H}{2}\right]\|\vg_2(\vw_k)\|^2-\frac{\alpha}{2}\|\nabla \gF(\vw_k)\|^2+\frac{\alpha}{2}\|\vg_1(\vw_k)-\vg_1(\vw_0)+\vg_1(\vw_0)\|^2\\
    \leq&\gF(\vw_k)-\left[\frac{\alpha}{2}-\frac{\alpha^2 H}{2}\right]\|\vg_2(\vw_k)\|^2-\frac{\alpha}{2}\|\nabla \gF(\vw_k)\|^2+\alpha\|\vg_1(\vw_k)-\vg_1(\vw_0)\|^2\\
    &+\alpha \|\vg_1(\vw_0)\|^2\\
    \overset{(b)}{\leq}&\gF(\vw_k)-\left[\frac{\alpha}{2}-\frac{\alpha^2 H}{2}\right]\|\vg_2(\vw_k)\|^2-\frac{\alpha}{2}\|\nabla \gF(\vw_k)\|^2+\frac{H^2\alpha^3}{4}\|\sum_{i=0}^{k-1} \vg_2(\vw_i)\|^2\\
    &+\alpha \|\vg_1(\vw_0)\|^2,
\end{align*}
\endgroup
where (a) is because $\nabla \gF(\vw_k)=\vg_1(\vw_k)+\vg_2(\vw_k)$, and (b) is because of the smoothness of $\gL$ and \cref{eq:update}.

Therefore, 
\begingroup
\allowdisplaybreaks
\begin{align*}
    &\min_{k} \|\nabla \gF(\vw_k)\|^2\\
    \leq& \frac{1}{K}\sum_{k=0}^{K-1} \|\nabla \gF(\vw_k)\|^2\\
    \leq& \frac{2}{\alpha K}\sum_{k=0}^{K-1}\left[\gF(\vw_k)-\gF(\vw_{k+1})+\frac{H^2\alpha^3}{4}\|\sum_{i=0}^{k-1} \vg_2(\vw_i)\|^2
    +\alpha \|\vg_1(\vw_0)\|^2-\left[\frac{\alpha}{2}-\frac{\alpha^2 H}{2}\right]\|\vg_2(\vw_k)\|^2\right]\\
    \leq& \frac{2}{\alpha K}[\gF(\vw_0)-\gF(\vw_{K})]+\frac{H^2\alpha^2}{2(K-1)}\sum_{k=1}^{K-1}\|\sum_{i=0}^{k-1} \vg_2(\vw_i)\|^2+2\|\vg_1(\vw_0)\|^2-\frac{1-\alpha H}{K}\sum_{k=0}^{K-1}\|\vg_2(\vw_k)\|^2\\
    \overset{(a)}{\leq}&\frac{2}{\alpha K}[\gF(\vw_0)-\gF(\vw_{K})]+\frac{\gamma^2}{2}\|\vg_1(\vw_0)\|^2+2\|\vg_1(\vw_0)\|^2-\frac{1-\alpha H}{K}\sum_{k=0}^{K-1}\|\vg_2(\vw_k)\|^2\\
    \leq& \frac{2}{\alpha K}[\gF(\vw_0)-\gF(\vw^*)]+\frac{4+\gamma^2}{2}\|\vg_1(\vw_0)\|^2-\frac{1-\alpha H}{K}\sum_{k=0}^{K-1}\|\vg_2(\vw_k)\|^2\\
    \leq& \frac{2}{\alpha K}[\gF(\vw_0)-\gF(\vw^*)]+\frac{4+\gamma^2}{2}\|\vg_1(\vw_0)\|^2
\end{align*}
\endgroup
where (a) holds due to $\gF(\vw^*)\leq\gF(\vw_{K})$ and $\|\sum_{i=0}^{k-1} \vg_2(\vw_i)\|^2\leq \frac{\gamma^2}{\alpha^2 H^2}\|\vg_1(\vw_0)\|^2$ based on \cref{sum}.

\section{Proof of Theorem \ref{thm:2}}

(1) For Rule $\#1$, we have
\begin{align}\label{gpm}
    \vw^c=\vw-\alpha [\vg_2(\vw)-\proj_{S^1}(\vg_2(\vw))]=\vw-\alpha\Tilde{\vg_2}(\vw).
\end{align}
For Rule $\#2$, we have
\begin{align}\label{eq:r3}
    \vw^r=\vw-\alpha \vg_2(\vw).
\end{align}

Based on the smoothness of the objective function, we can have an upper bound on $\gF(\vw^r)$:
\begin{align}\label{upper}
    \gF(\vw^r)\leq \gF(\vw)-\left[\alpha-\frac{\alpha^2 H}{2}\right]\|\vg_2(\vw)\|^2-\alpha\langle \vg_1(\vw),\vg_2(\vw) \rangle,
\end{align}
and a lower bound on $\gF(\vw^c)$:
\begin{align}\label{lower}
    \gF(\vw^c)\geq \gF(\vw)+\langle \nabla \gF(\vw), \vw^c-\vw\rangle - \frac{H}{2}\|\vw^c-\vw\|^2.
\end{align}

Combining \cref{upper} and \cref{lower}, it can be shown that
\begingroup
\allowdisplaybreaks
\begin{align}\label{eq:thm2}
    &\gF(\vw^r)\nonumber\\
    \leq&\gF(\vw^c)-\langle \nabla \gF(\vw), \vw^c-\vw\rangle + \frac{H}{2}\|\vw^c-\vw\|^2-\left[\alpha-\frac{\alpha^2 H}{2}\right]\|\vg_2(\vw)\|^2-\alpha\langle \vg_1(\vw),\vg_2(\vw) \rangle\nonumber\\
    =&\gF(\vw^c)-\langle \vg_1(\vw)+\vg_2(\vw), -\alpha\Tilde{\vg_2}(\vw)\rangle + \frac{H\alpha^2}{2}\|\Tilde{\vg_2}(\vw)\|^2-\left[\alpha-\frac{\alpha^2 H}{2}\right]\|\vg_2(\vw)\|^2\nonumber\\
    &-\alpha\langle \vg_1(\vw),\vg_2(\vw) \rangle\nonumber\\
    =& \gF(\vw^c)+\alpha\langle \vg_1(\vw), \Tilde{\vg_2}(\vw)\rangle+ \alpha\langle \vg_2(\vw), \Tilde{\vg_2}(\vw)\rangle+ \frac{H\alpha^2}{2}\|\Tilde{\vg_2}(\vw)\|^2-\left[\alpha-\frac{\alpha^2 H}{2}\right]\|\vg_2(\vw)\|^2\nonumber\\
    &-\alpha\langle \vg_1(\vw),\vg_2(\vw) \rangle\nonumber\\
    =&\gF(\vw^c)+\alpha\|\Tilde{\vg_2}(\vw)\|^2+ \frac{H\alpha^2}{2}\|\Tilde{\vg_2}(\vw)\|^2-\left[\alpha-\frac{\alpha^2 H}{2}\right]\|\vg_2(\vw)\|^2
    -\alpha\langle \vg_1(\vw),\vg_2(\vw) \rangle
    \end{align}
\endgroup
where the last equality is true because both $\vg_1(\vw)$ and $\proj_1(\vg_2(\vw))$ are orthogonal to $\Tilde{\vg_2}(\vw)$.

For the term $\langle \vg_1(\vw),\vg_2(\vw) \rangle$, based on \cref{eq:inner}, it follows that
\begin{align}\label{eq:lowerinner}
    &\langle \vg_1(\vw),\vg_2(\vw) \rangle\nonumber\\
    \geq& -\frac{H^2}{4}\|\vw-\vw_0\|^2-\frac{1}{2}\|\vg_2(\vw)\|^2-\frac{1}{2}\|\vg_1(\vw_0)\|^2+\langle \vg_1(\vw_0),\vg_2(\vw_0) \rangle\nonumber\\
    \geq& -\frac{H^2}{4}\|\vw-\vw_0\|^2-\frac{1}{2}\|\vg_2(\vw)\|^2-\frac{1}{2}\|\vg_1(\vw_0)\|^2+\epsilon_2 \|\vg_1(\vw_0)\|\|\vg_2(\vw_0)\|.
\end{align}

Suppose that $\vw$ is the model update at $n$-th iteration where $n\leq K$. For Rule $\#1$,
\begin{align*}
    \|\vw-\vw_0\|^2=& \alpha^2 \|\sum_{i=0}^n \Tilde{\vg_2}(\vw_i)\|^2\\
    \leq& \frac{\gamma^2 \|\vg_1(\vw_0)\|^2}{H^2B^2K^2}n\sum_{i=0}^n\|\Tilde{\vg_2}(\vw_i)\|^2\\
    \leq&\frac{\gamma^2 n^2 \|\vg_1(\vw_0)\|^2}{H^2K^2}\\
    \leq&\frac{\gamma^2 \|\vg_1(\vw_0)\|^2}{H^2}.
\end{align*}
Similarly for Rule $\#2$, we can also obtain that 
\begin{align*}
    \|\vw-\vw_0\|^2\leq&\frac{\gamma^2 \|\vg_1(\vw_0)\|^2}{H^2}.
\end{align*}

Therefore, continuing with \cref{eq:lowerinner}, we can have
\begin{align*}
     &\langle \vg_1(\vw),\vg_2(\vw) \rangle\\
     \geq& -\frac{2+\gamma^2 }{4}\|\vg_1(\vw_0)\|^2+\epsilon_2 \|\vg_1(\vw_0)\|\|\vg_2(\vw_0)\|-\frac{1}{2}\|\vg_2(\vw)\|^2\\
     \geq&-\frac{1}{2}\|\vg_2(\vw)\|^2
\end{align*}
where the last inequality holds because $\epsilon_2\geq \frac{(2+\gamma^2)\|\vg_1(\vw_0)\|}{4\|\vg_2(\vw_0)\|}$.

Based on \cref{eq:thm2}, it follows that
\begingroup
\allowdisplaybreaks
\begin{align*}
    \gF(\vw^r)\leq& \gF(\vw^c)+\alpha\|\Tilde{\vg_2}(\vw)\|^2+ \frac{H\alpha^2}{2}\|\Tilde{\vg_2}(\vw)\|^2-\left[\alpha-\frac{\alpha^2 H}{2}\right]\|\vg_2(\vw)\|^2+\frac{\alpha}{2}\|\vg_2(\vw)\|^2\\
    =& \gF(\vw^c)-\left[\frac{\alpha}{2}-\frac{\alpha^2 H}{2}\right]\|\vg_2(\vw)\|^2+\left[\alpha+\frac{\alpha^2 H}{2}\right]\|\Tilde{\vg_2}(\vw)\|^2\\
    \overset{(a)}{\leq}&\gF(\vw^c)-\left[\frac{\alpha}{2}-\frac{\alpha^2 H}{2}\right]\|\vg_2(\vw)\|^2+(1-\epsilon_1^2)\left[\alpha+\frac{\alpha^2 H}{2}\right]\|\vg_2(\vw)\|^2\\
    \overset{(b)}{\leq} &\gF(\vw^c)
\end{align*}
\endgroup

where (a) holds because 
\begin{align*}
    \|\vg_2(\vw)\|^2=&\|\proj_1(\vg_2(\vw))+\Tilde{\vg_2}(\vw)\|^2\\
    =&\|\proj_1(\vg_2(\vw))\|^2+\|\Tilde{\vg_2}(\vw)\|^2\\
    \geq& \epsilon_1^2 \|\vg_2(\vw)\|^2+\|\Tilde{\vg_2}(\vw)\|^2,
\end{align*}
and (b) is true  because $\epsilon_1\geq \sqrt{\frac{1+2\alpha H}{2+\alpha H}}$.

(2) It can be seen that
\begin{align*}
    \gL_1(\vw_k)\leq & \gL_1(\vw_0)+\langle\vg_1(\vw_0), \vw_k-\vw_0\rangle+\frac{H}{4}\|\vw_k-\vw_0\|^2 \\
    =& \gL_1(\vw_0)+\langle\vg_1(\vw_0),-\alpha \sum_{i=0}^{k-1} \vg_2(\vw_i)\rangle+\frac{\alpha^2 H}{4}\|\sum_{i=0}^{k-1} \vg_2(\vw_i)\|^2\\
    =&\gL_1(\vw_0)-\alpha \sum_{i=0}^{k-1}\langle \vg_1(\vw_0),\vg_2(\vw_i) \rangle +\frac{\alpha^2 H}{4}\|\sum_{i=0}^{k-1}\vg_2(\vw_i)\|^2\\
    \leq& \gL_1(\vw_0)-\alpha\epsilon_2\|\vg_1(\vw_0)\|[\sum_{i=0}^{k-1}\|\vg_2(\vw_i)\|]+\frac{\alpha^2 Hk}{4}\sum_{i=0}^{k-1}\|\vg_2(\vw_i)\|^2.
\end{align*}

For $\alpha\leq \frac{4\epsilon_2\|\vg_1(\vw_0)\|}{HBk^{1.5}}$, we can have
\begin{align*}
    \frac{\alpha Hk}{4}\sum_{i=0}^{k-1}\|\vg_2(\vw_i)\|^2
    \leq& \frac{\epsilon_2 \|\vg_1(\vw_0)\|}{B\sqrt{k}}\sum_{i=0}^{k-1}\|\vg_2(\vw_i)\|^2\\
    \leq& \frac{\epsilon_2 \|\vg_1(\vw_0)\| \left(\sum_{i=0}^{k-1}\|\vg_2(\vw_i)\|^2\right)}{\sqrt{\sum_{i=0}^{k-1}\|\vg_2(\vw_i)\|^2}}\\
    \leq& \epsilon_2 \|\vg_1(\vw_0)\| \sqrt{\sum_{i=0}^{k-1}\|\vg_2(\vw_i)\|^2}\\
    \leq& \epsilon_2 \|\vg_1(\vw_0)\|[\sum_{i=0}^{k-1}\|\vg_2(\vw_i)\|].
\end{align*}

Therefore, it follows that $\gL_1(\vw_k)\leq\gL_1(\vw_0)$.

\section{More experimental results}

\begin{figure*}[ht]
    \centering
    \includegraphics[width=0.31\linewidth]{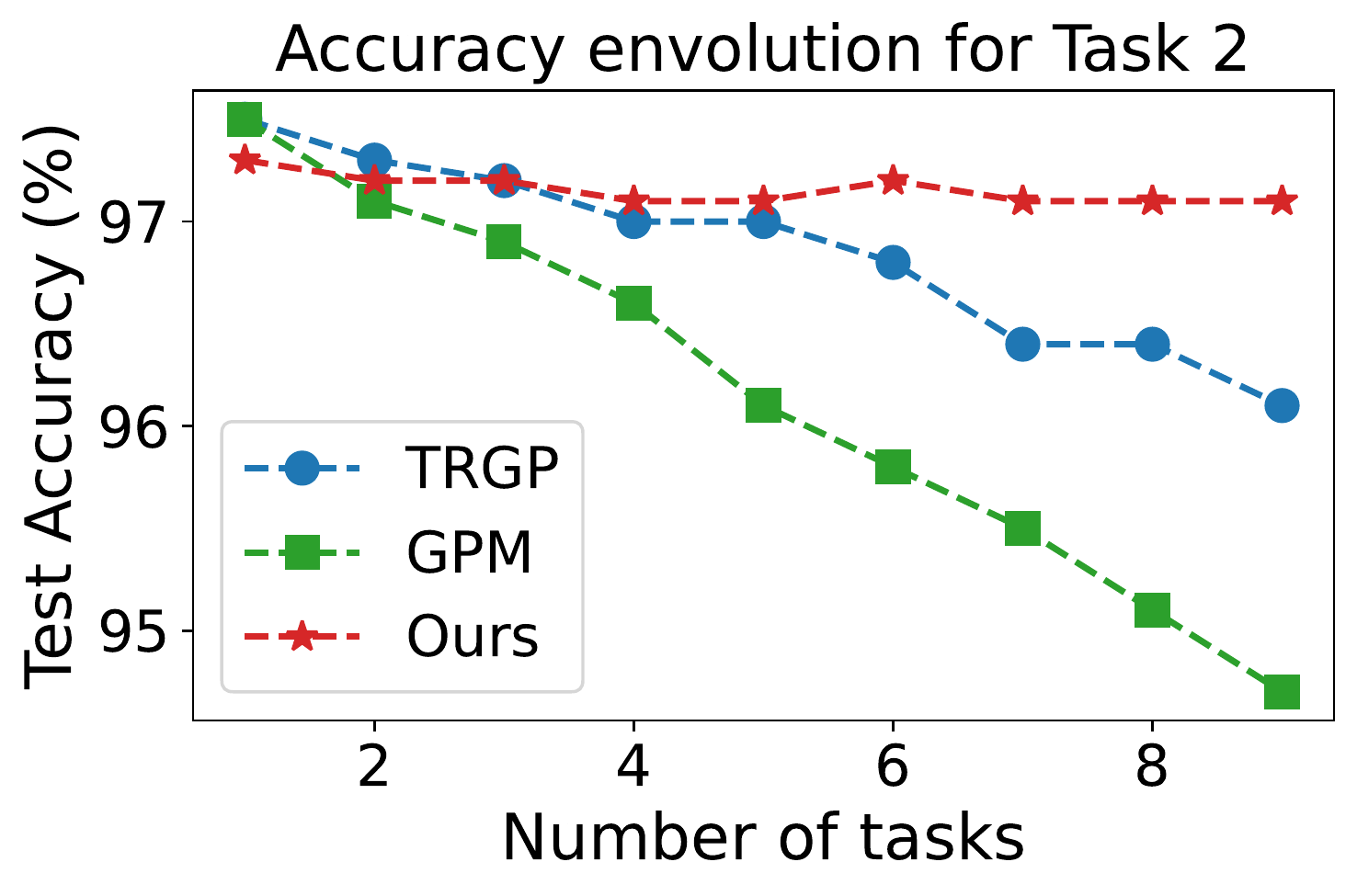} 
    \includegraphics[width=0.31\linewidth]{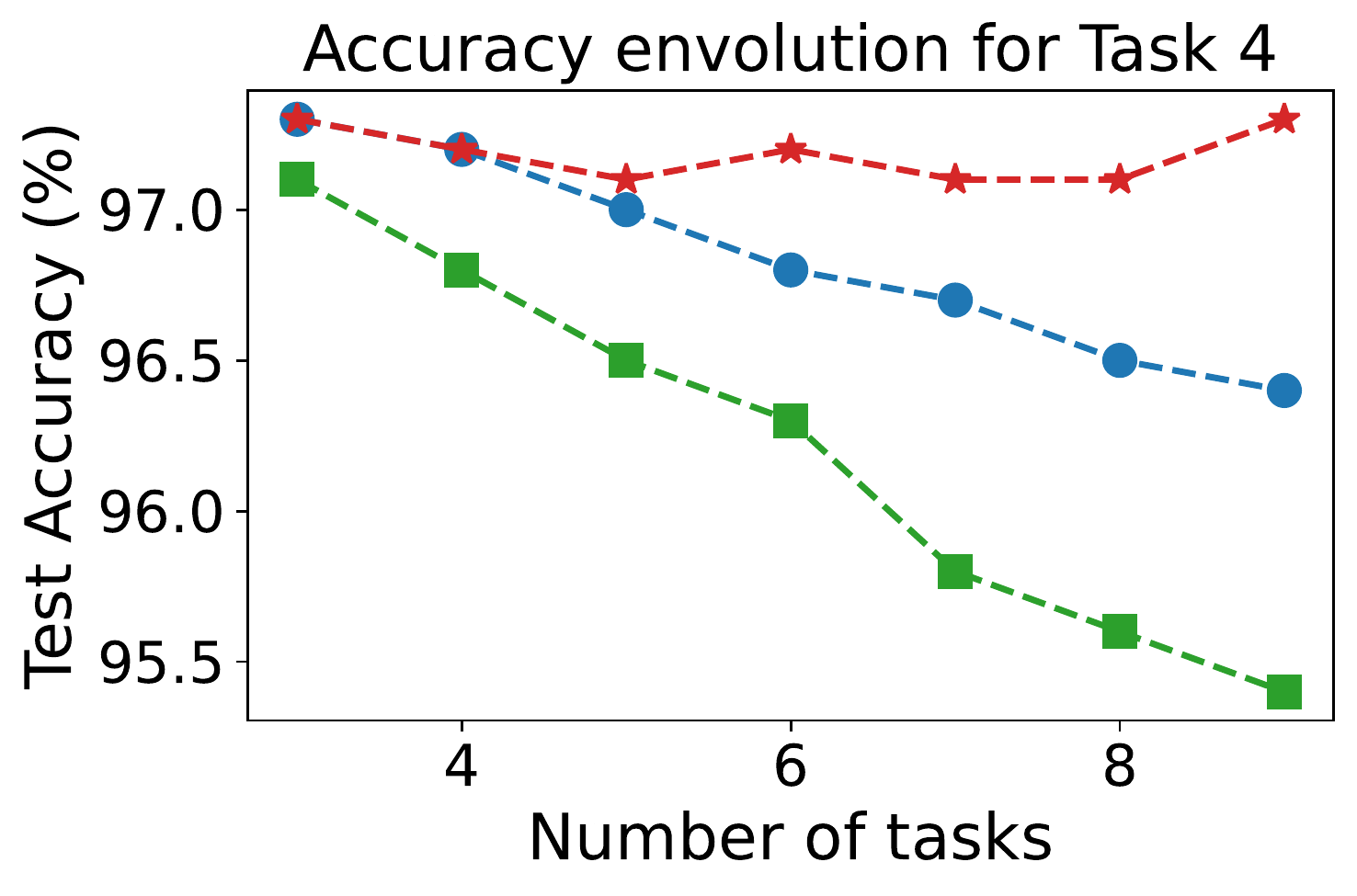} 
    \includegraphics[width=0.31\linewidth]{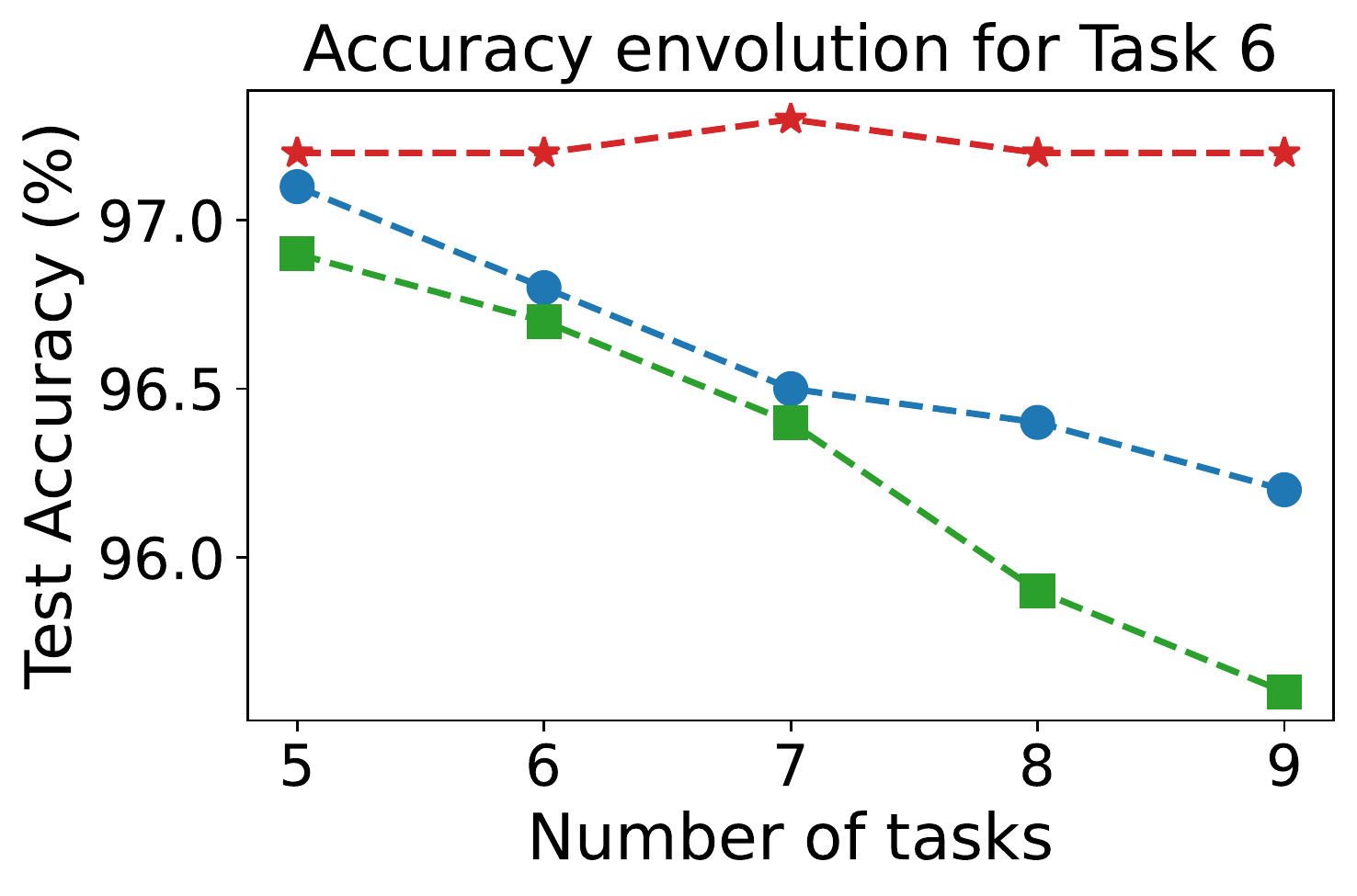}
    \\ (a) PMNIST\\
    \includegraphics[width=0.31\linewidth]{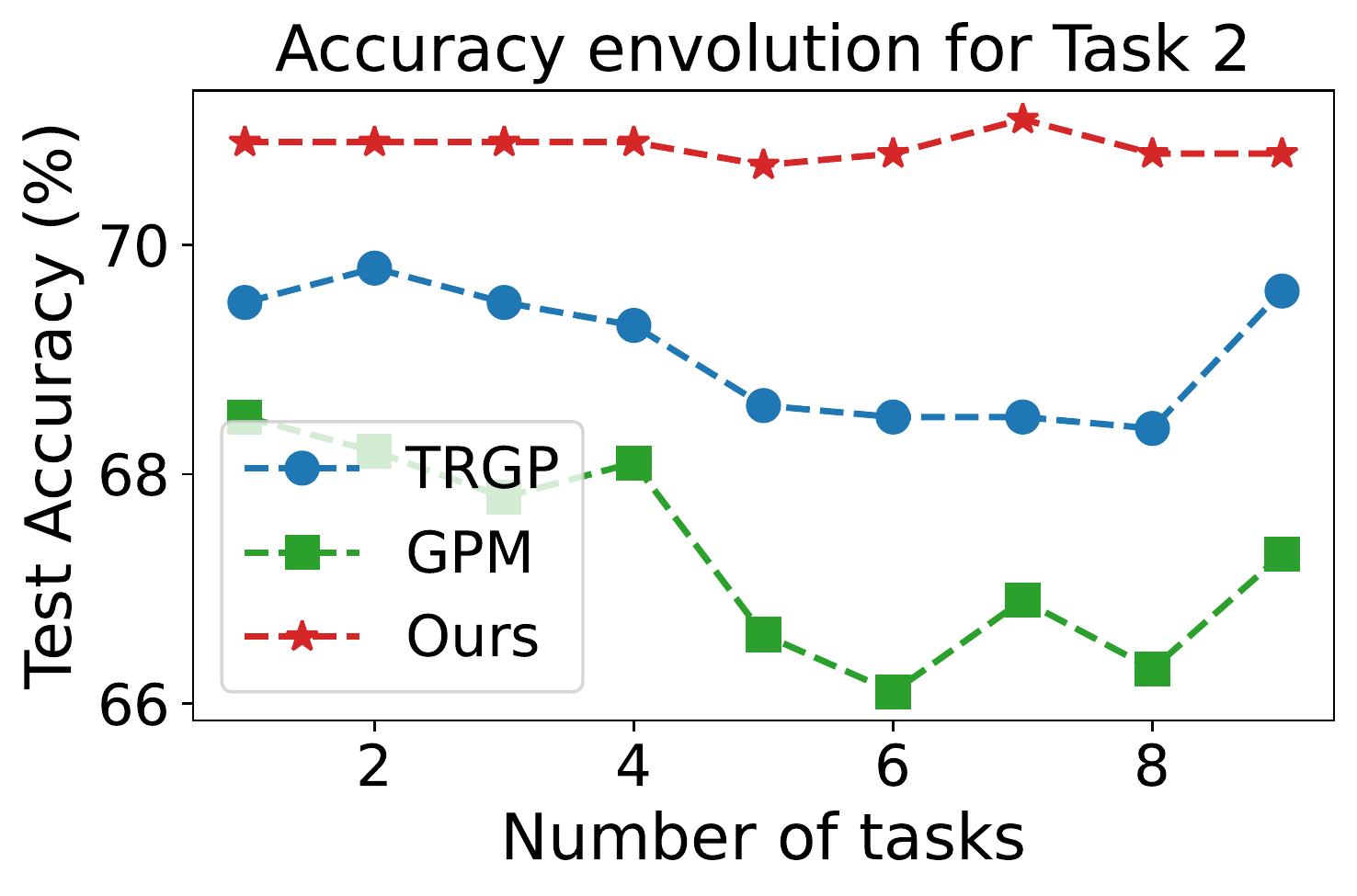} 
    \includegraphics[width=0.31\linewidth]{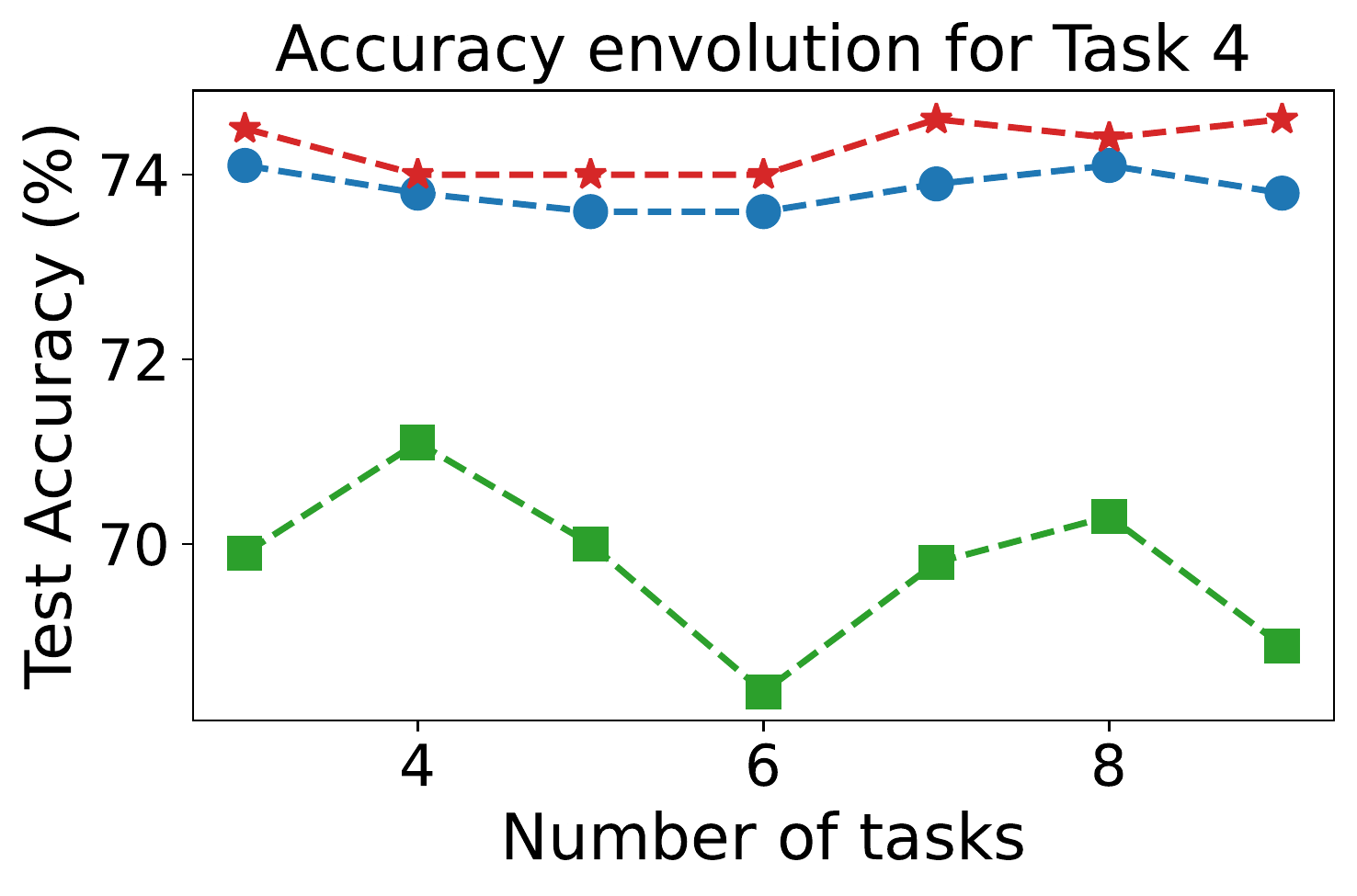} 
    \includegraphics[width=0.31\linewidth]{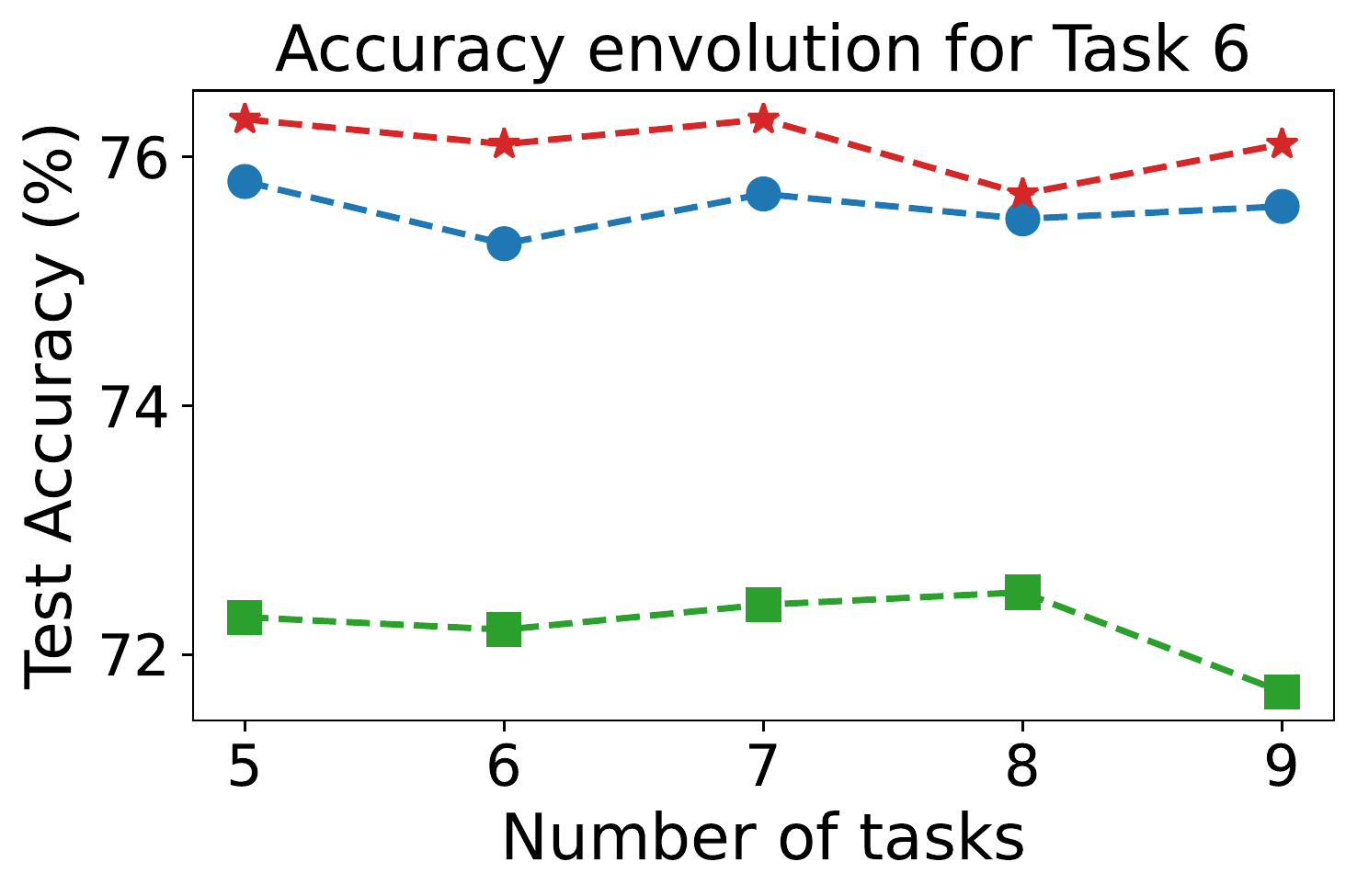} \\  (b) CIFAR-100 Split\\   
    \includegraphics[width=0.31\linewidth]{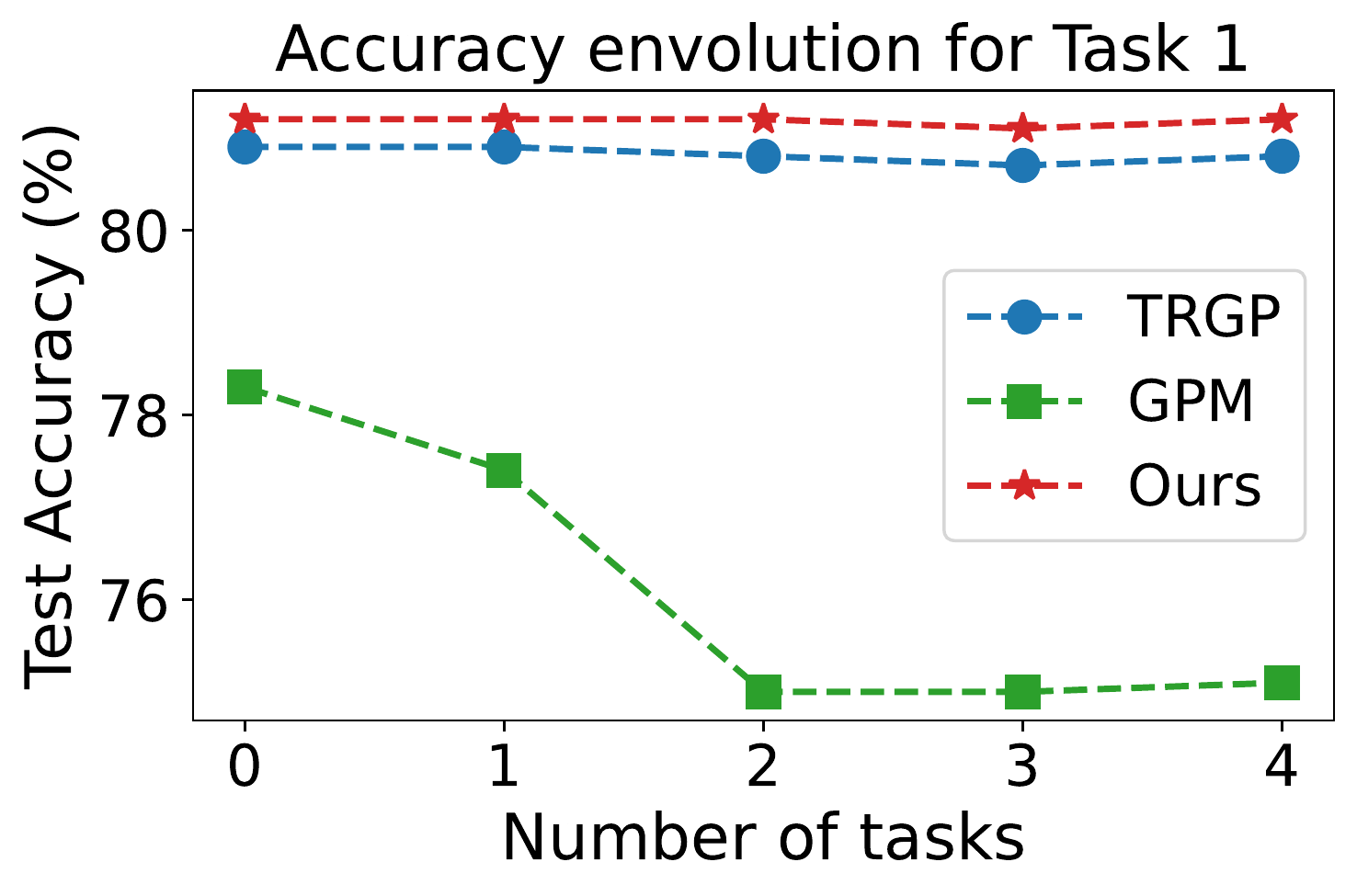} 
    \includegraphics[width=0.31\linewidth]{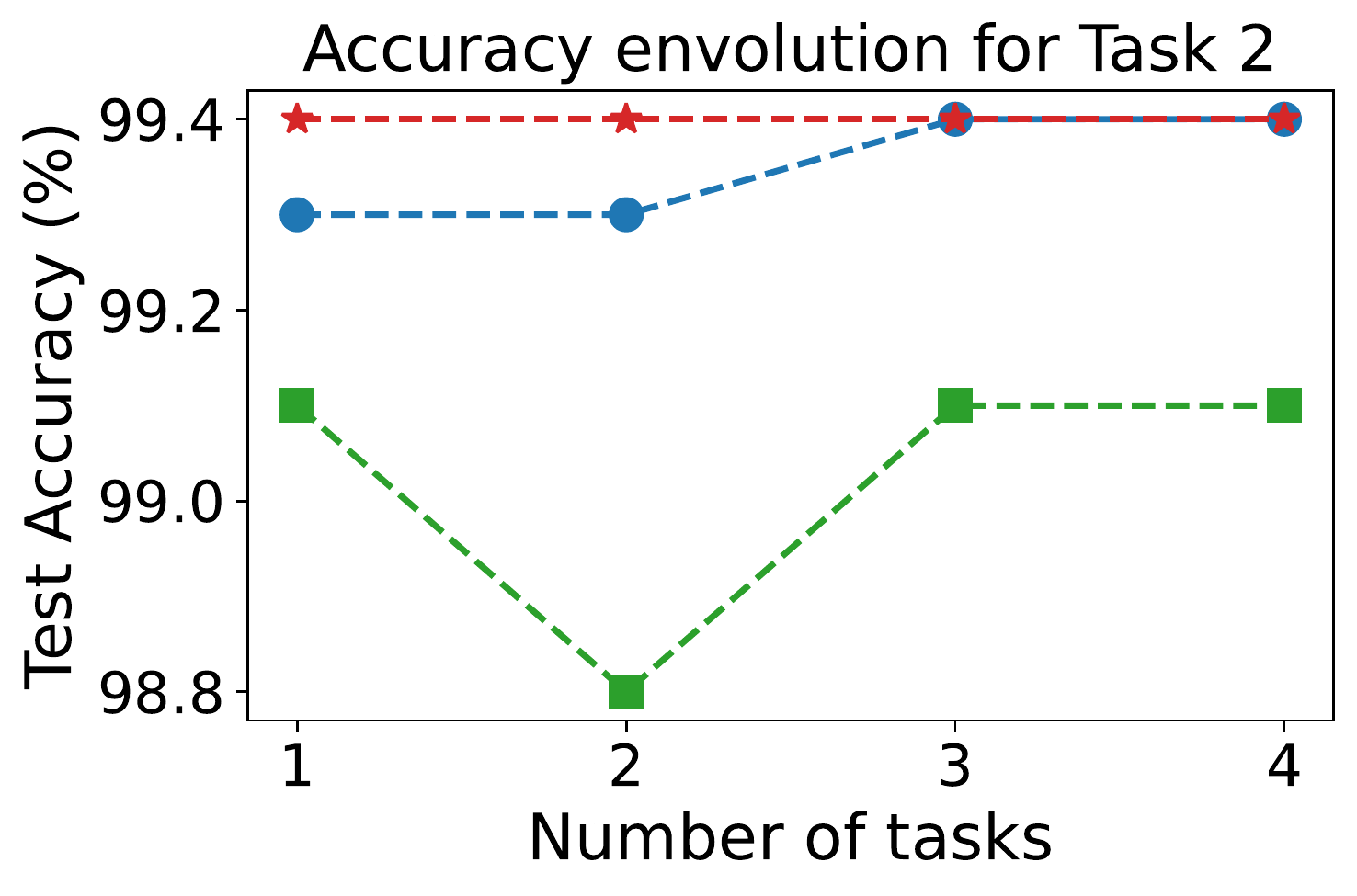} 
    \includegraphics[width=0.31\linewidth]{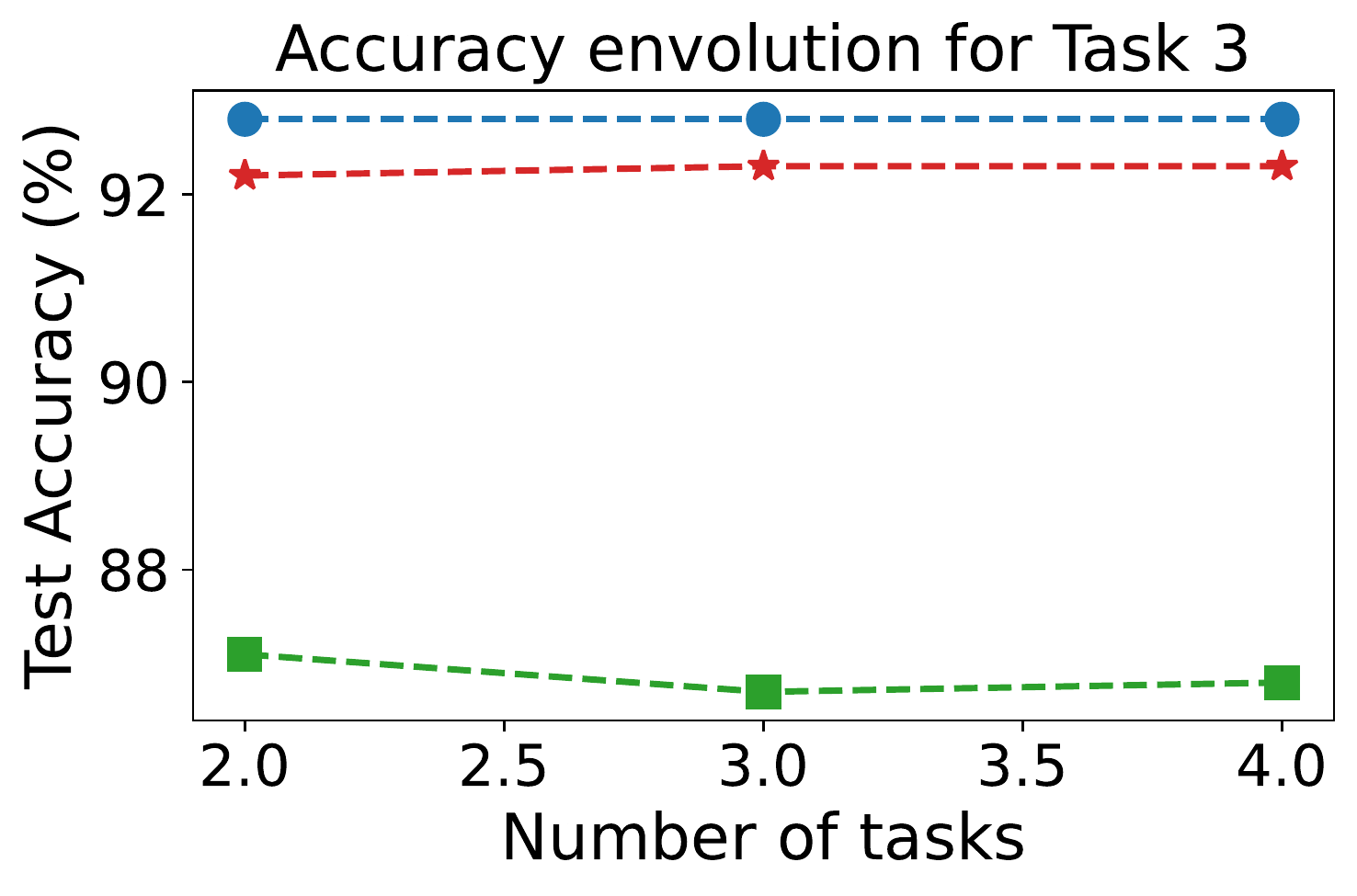} \\  (c) 5-Dataset\\ 
    \includegraphics[width=0.31\linewidth]{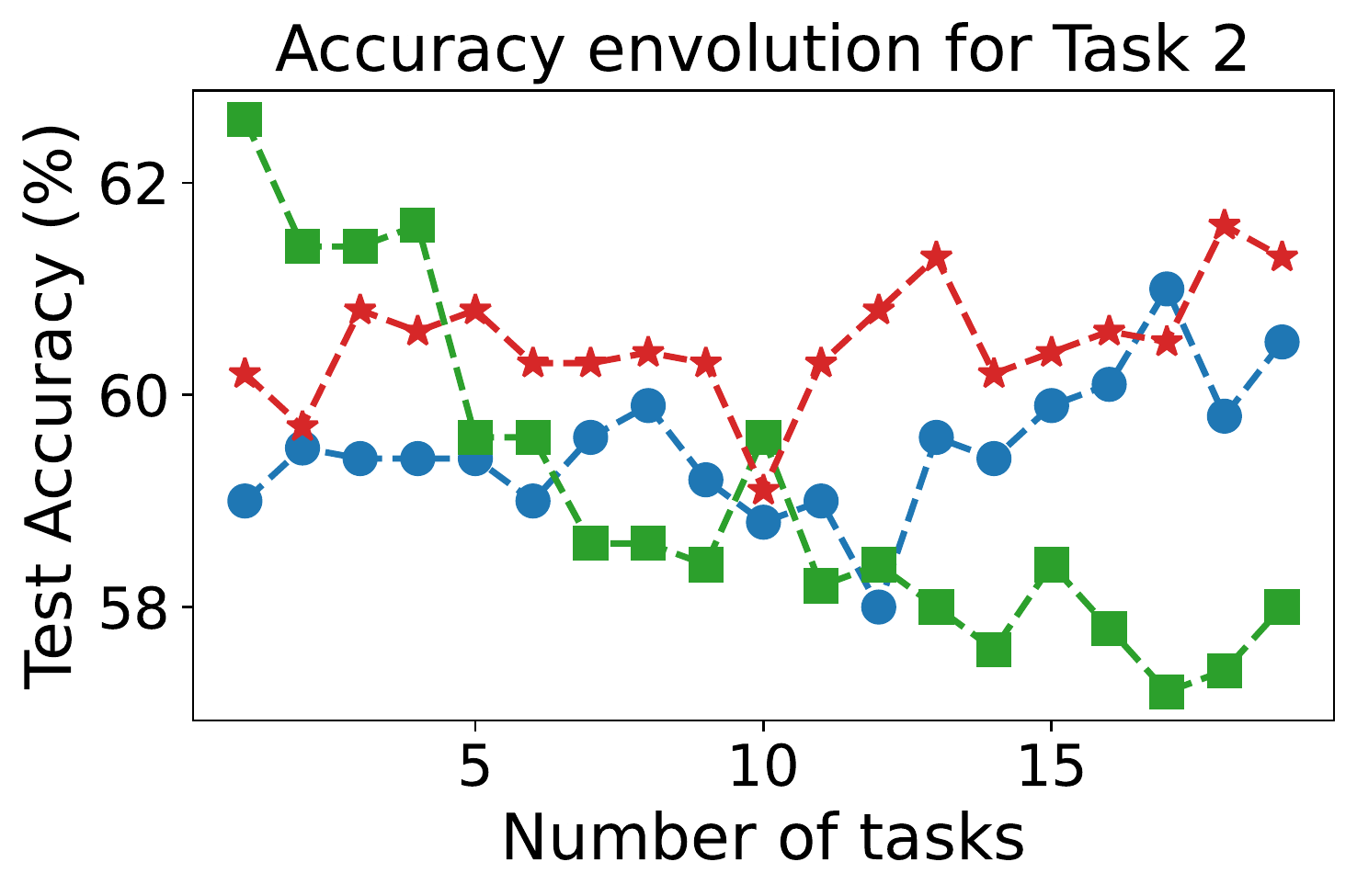} 
    \includegraphics[width=0.31\linewidth]{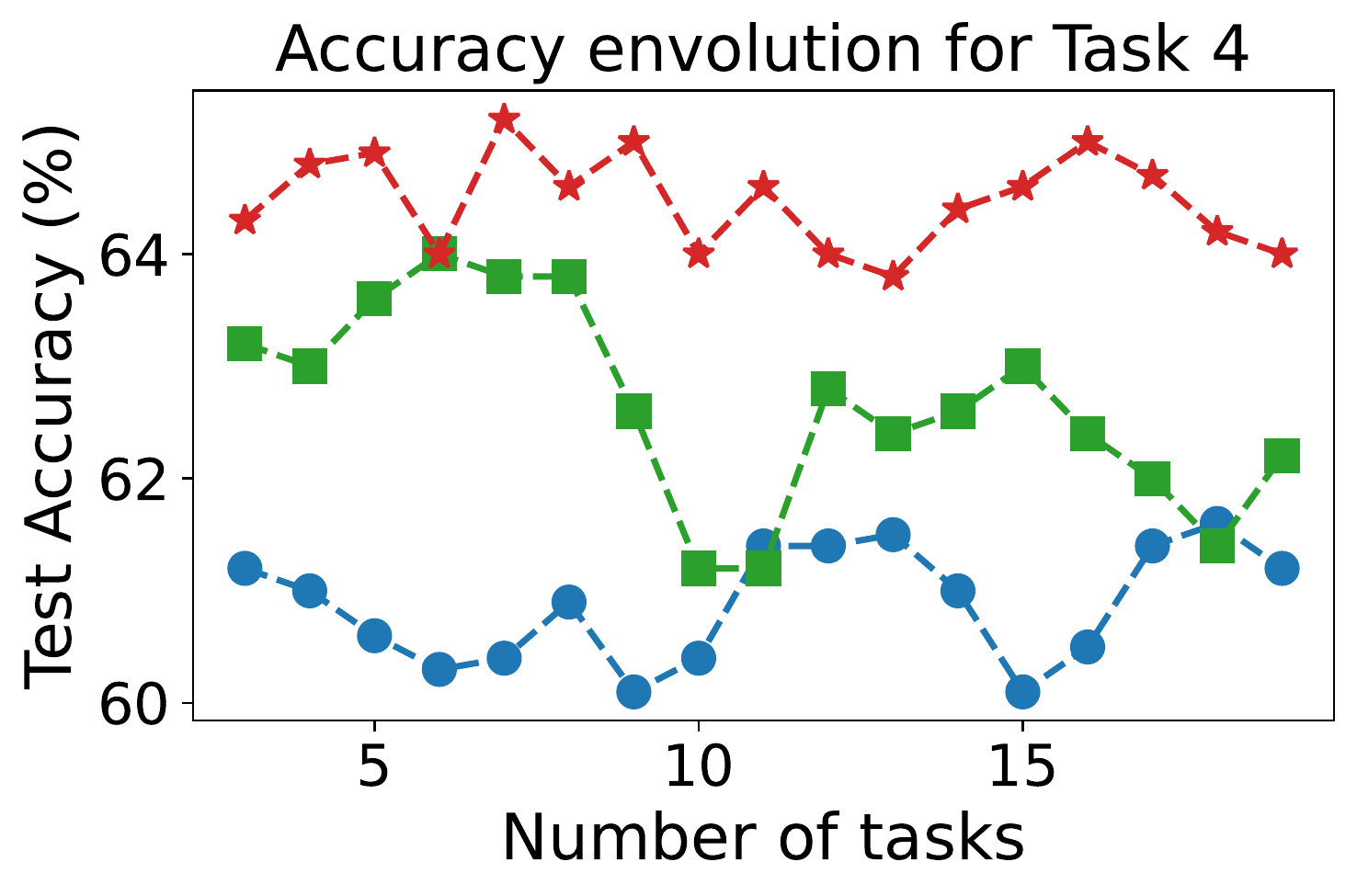} 
    \includegraphics[width=0.31\linewidth]{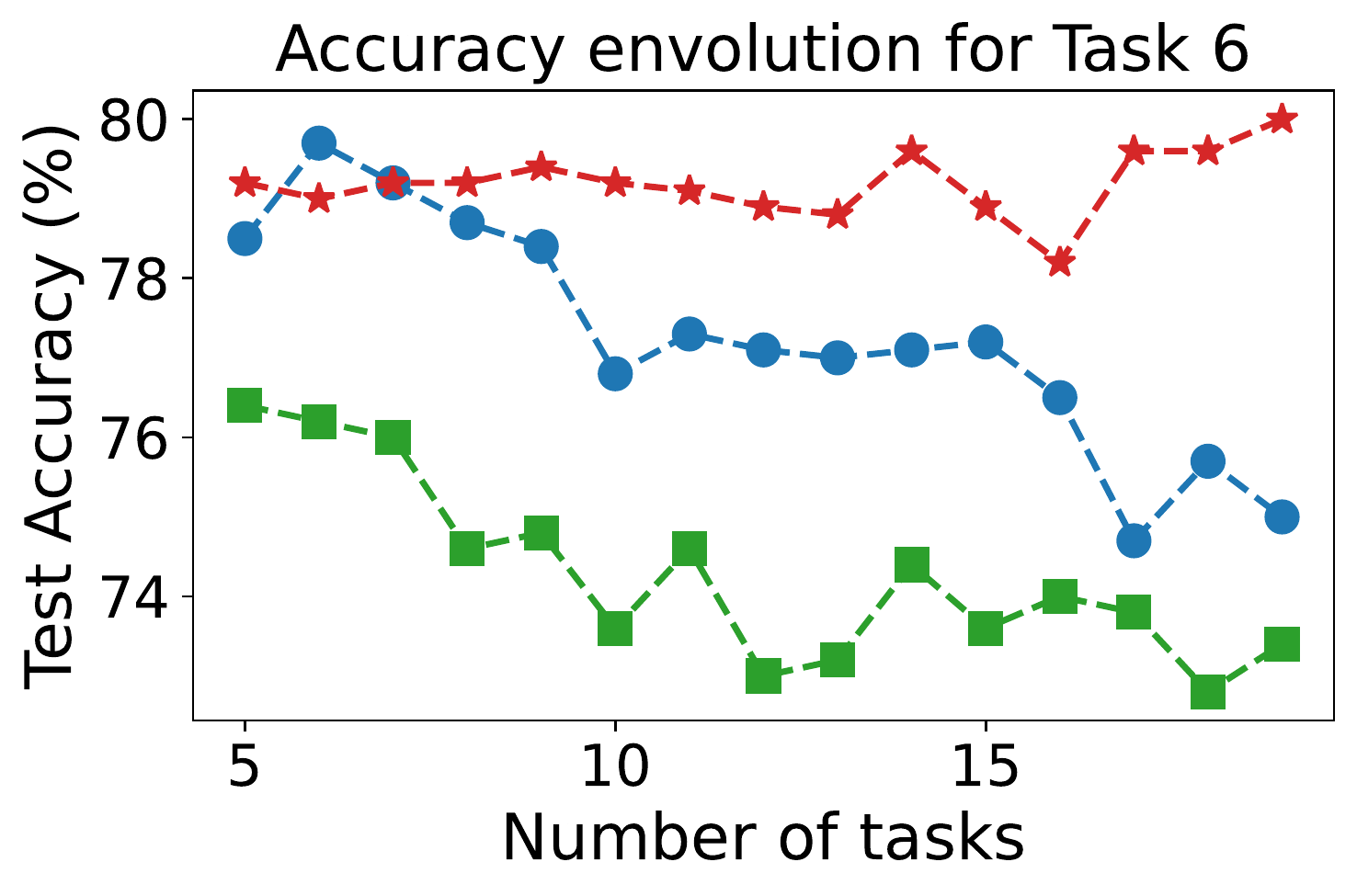} \\  (d) MiniImageNet\\
    \caption{Accuracy evolution for different tasks on PMNIST, CIFAR-100 Split, 5-Datasets and MiniImageNet, respectively.} 
    \label{fig:curve}
\end{figure*}

We also study the accuracy evolution curves for each task during the continual learning procedure in Figure \ref{fig:curve}. To clearly demonstrate the accuracy evolution behaviors, we selectively study task 2, task 4, task 6 in PMNIST \cite{lecun1998mnist}, CIFAR-100 Split \cite{krizhevsky2009learning}, MiniImageNet \cite{vinyals2016matching}, and task 1, task 2, task 3 in 5-Dataset, by comparing the performance among CUBER, TRGP and GPM. As shown in Figure \ref{fig:curve}, CUBER (the red curve) has the most stable performance on most tasks, as a result of the better backward knowledge transfer facilitated by selectively updating the learnt model of old tasks. 
All the experiments are conducted by using one Nvidia Quadro RTX 5000 GPU. 

\section{More experimental details}

\subsection{Network details}

\textbf{3-layer fully-connected network:} The network consists of 3 fully-connected layers with 784, 100, 100 units respectively. We use ReLU activation layer after the first two layers.

\textbf{5-layer AlexNet:} Following \cite{saha2021gradient,lin2022trgp}, the AlexNet used for experiments on Split CIFAR-100 consists of 3 convolutional layers and 2 fully-connected layers, where batch normalization is added in each layer expect the classifier layer. The convolutional layers have 64, 128 and 256 filters with $4\times 4$, $3\times 3$ and $2\times 2$ kernel sizes, respectively, and each fully-connected layer contains 2048 units. We use ReLU activation function and $2\times 2$ max-pooling after the convolutional layers, and dropout of 0.2 for the first two layers and 0.5 for other layers.

\textbf{A reduced ResNet-18:} Following \cite{lopez2017gradient, saha2021gradient}, we use a reduced ResNet-18 for experiments on both 5-Datasets and miniImageNet. It includes 17 convolutional layers with 3 short-cut connections. The convolutional layers in 4 stages have 180, 360, 720, 1440 filters, respectively. We adapt the 2x2 average pooling layer before the classifier. In addition, we use convolution with stride 2 in the first layer for miniImageNet, since it has larger input resolution (i.e., 84x84) than 5-dataset (i.e., 32x32).

\subsection{List of hyperparameters}

In what follows, we list the hyperparameters for all the methods considered in this work. For TRGP, we use the hyperparameters provided in \cite{lin2022trgp}. For other baseline methods, we use the hyperparameters for GPM as provided in \cite{saha2021gradient}, and provide the hyperparameters for the rest following \cite{saha2021gradient} for being consistent with the corresponding reported results herein.

As shown in the Table \ref{tab:hp}, we use `lr' to represent the initial learning rate, and `cifar', `mini', `5d' and `pm' to represent `Split CIFAR-100', `Split MiniImageNet', `5-Dataset' and `Permuted MNIST', respectively. 

\begin{table*}[!htbp]
\centering
\caption{List of hyperparameters for CUBER and the related baseline methods.}
\label{tab:hp}
\begin{tabular}{cc}
\toprule
Methods         & Hyperparameters               \\\midrule 
OWM & lr: 0.01 (cifar), 0.3 (pm)\\ \midrule
\multirow{2}{*}{EWC} &  lr: 0.05 (cifar), 0.03 (mini, 5d, pm)\\
& regularization coefficient ($\lambda$): 5000 (cifar, mini, 5d), 1000 (pm)\\ \midrule
\multirow{3}{*}{HAT} & lr: 0.05 (cifar), 0.03 (mini), 0.1 (5d) \\
& $s_{max}$: 400 (cifar, mini, 5d)\\
& $c$: 0.75 (cifar, mini, 5d) \\ \midrule
\multirow{2}{*}{A-GEM} & lr: 0.05 (cifar), 0.1 (mini, 5d, pm) \\
& memory size (number of samples): 2000 (cifar), 500 (mini), 3000 (5d), 1000 (pm)\\ \midrule
\multirow{2}{*}{ER$\_$Res} & lr: 0.05 (cifar), 0.1 (mini, 5d, pm)\\
& memory size (number of samples): 2000 (cifar), 500 (mini), 3000 (5d), 1000 (pm) \\ \midrule
\multirow{4}{*}{GPM} & lr: 0.01 (cifar, pm), 0.1 (mini, 5d)\\
& number of samples for base extraction ($n$): 125 (cifar), 100 (mini, 5d), 300 (pm)\\
& $\epsilon_{th}$: 0.97, increase by 0.003 with $t$ (cifar); 0.985, increase by 0.003 with $t$ (mini) \\
& $\epsilon_{th}$: 0.965 (5d); 0.95 for the first layer, otherwise 0.99 (pm) \\\midrule
\multirow{5}{*}{TRGP} & lr: 0.01 (cifar, pm), 0.1 (mini, 5d) \\
& number of samples for base extraction ($n$): 125 (cifar), 100 (mini, 5d), 300 (pm)\\
& $\epsilon_{th}$: same with GPM \\
& $\epsilon^l$: 0.5 (cifar, mini, 5d, pm) \\\midrule
\multirow{5}{*}{CUBER} & lr: 0.01 (cifar, pm), 0.1 (mini, 5d)\\
& number of samples for base extraction ($n$): 125 (cifar), 100 (mini, 5d), 300 (pm)\\
& $\epsilon_{th}$:  same with GPM \\
& $\epsilon_1$: 0.5 (cifar, mini, 5d, pm); ~ $\epsilon_2$: 0 (cifar, mini, 5d, pm); $\lambda$: 1 (cifar, mini, 5d, pm)
 \\\bottomrule
\end{tabular}
\end{table*}

\textbf{Optimizer/learning rate:} For all experiments, we adapt the SGD optimizer by modifying the gradient based on the optimization problem (3). This is consistent with the orthogonal-projection based methods (e.g., \cite{farajtabar2020orthogonal, saha2021gradient, lin2022trgp}) where the gradient direction used in the plain SGD optimizer is modified to minimize the interference to the old tasks.
The learning rates for different datasets are shown in Table \ref{tab:hp}. Here for PMNIST, we use a fixed learning rate, while for other datasets the learning rate decays during the training process.

\textbf{Early stopping:}
For all experiments, the minimum learning rate is set to 1e-5, the learning rate decay factor is set to 2, and the number of holds before decaying the learning rate is 6. After each model update, we evaluate the validation loss for the current task using its validation dataset, and count the times whenever the validation loss increases. When the counter is greater than 6, we decay the learning rate by 2 and reset the counter to 0. The training process will be stopped when the learning rate decays to the minimum learning rate.

\subsection{Memory and time cost}

\textbf{Storing gradient:} After learning each task, most elements in the gradient matrix for each layer can be close to zero. When evaluating the conditions for regime 3, we flatten the gradient matrix into a vector for each layer, e.g., a gradient matrix in $\mathbb{R}^{m\times n}$ to a gradient vector in $\mathbb{R}^{mn}$. The inner product between gradient matrices hence becomes the inner product between gradient vectors, where the near-zero elements have little impact on the value of the product. Therefore, to save the computation cost and memory, we prune the layer-wise average gradient vector after learning each task with a certain sparsity ratio, and only store a gradient vector with non-zero elements and the corresponding indices in the original gradient vector.

\begin{figure*}[ht]
    \centering
    \includegraphics[width=0.21\linewidth]{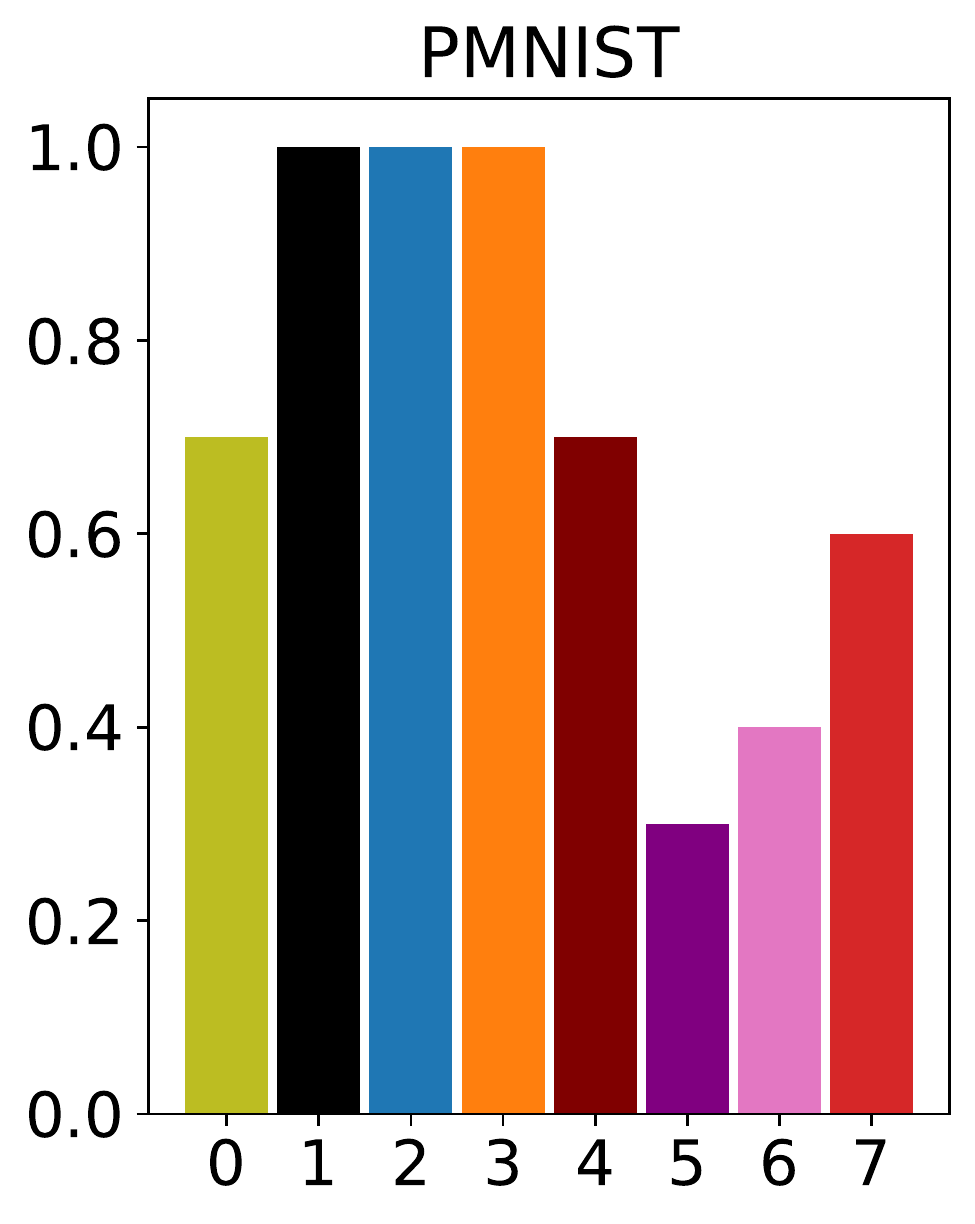} 
    \includegraphics[width=0.21\linewidth]{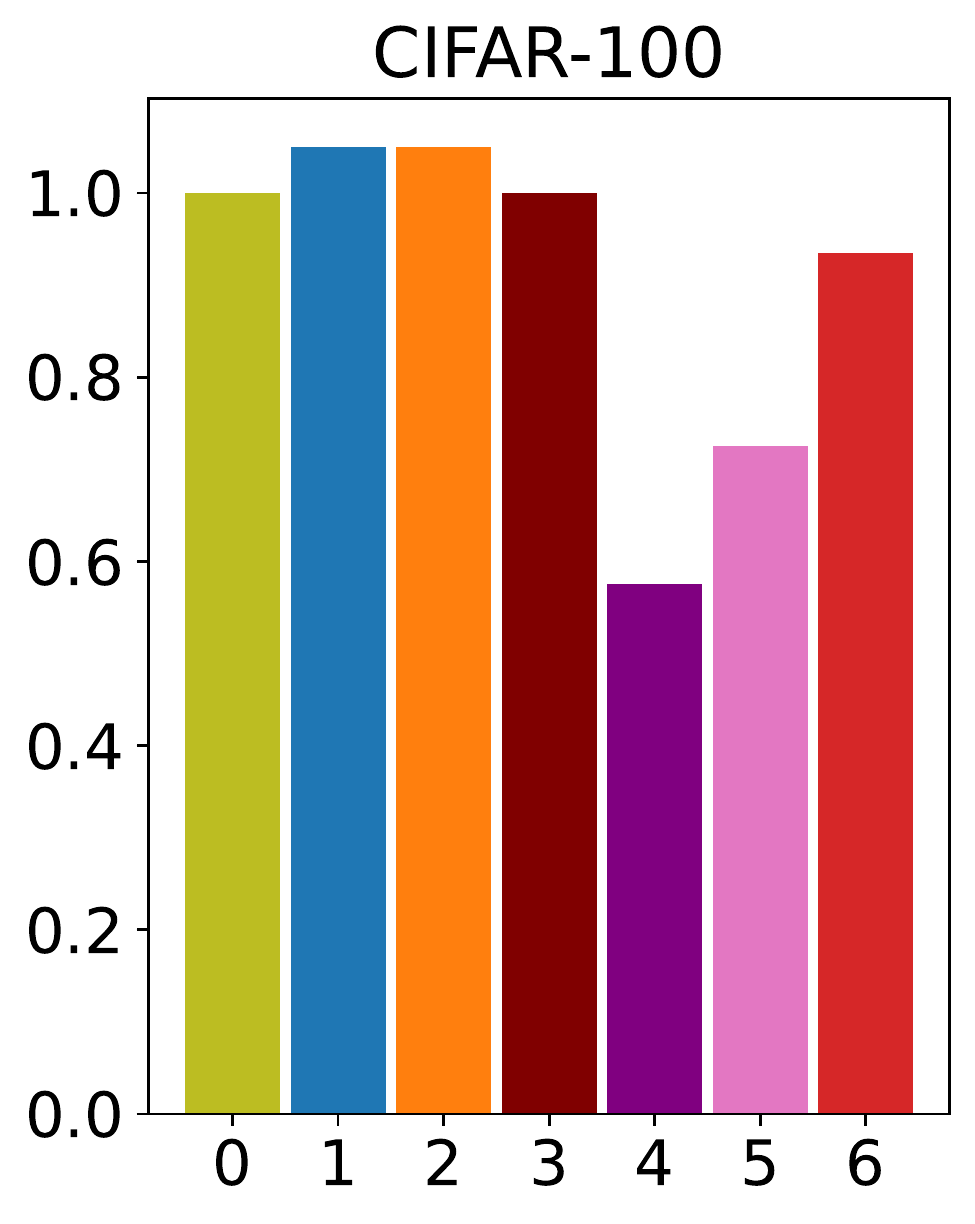} 
    \includegraphics[width=0.21\linewidth]{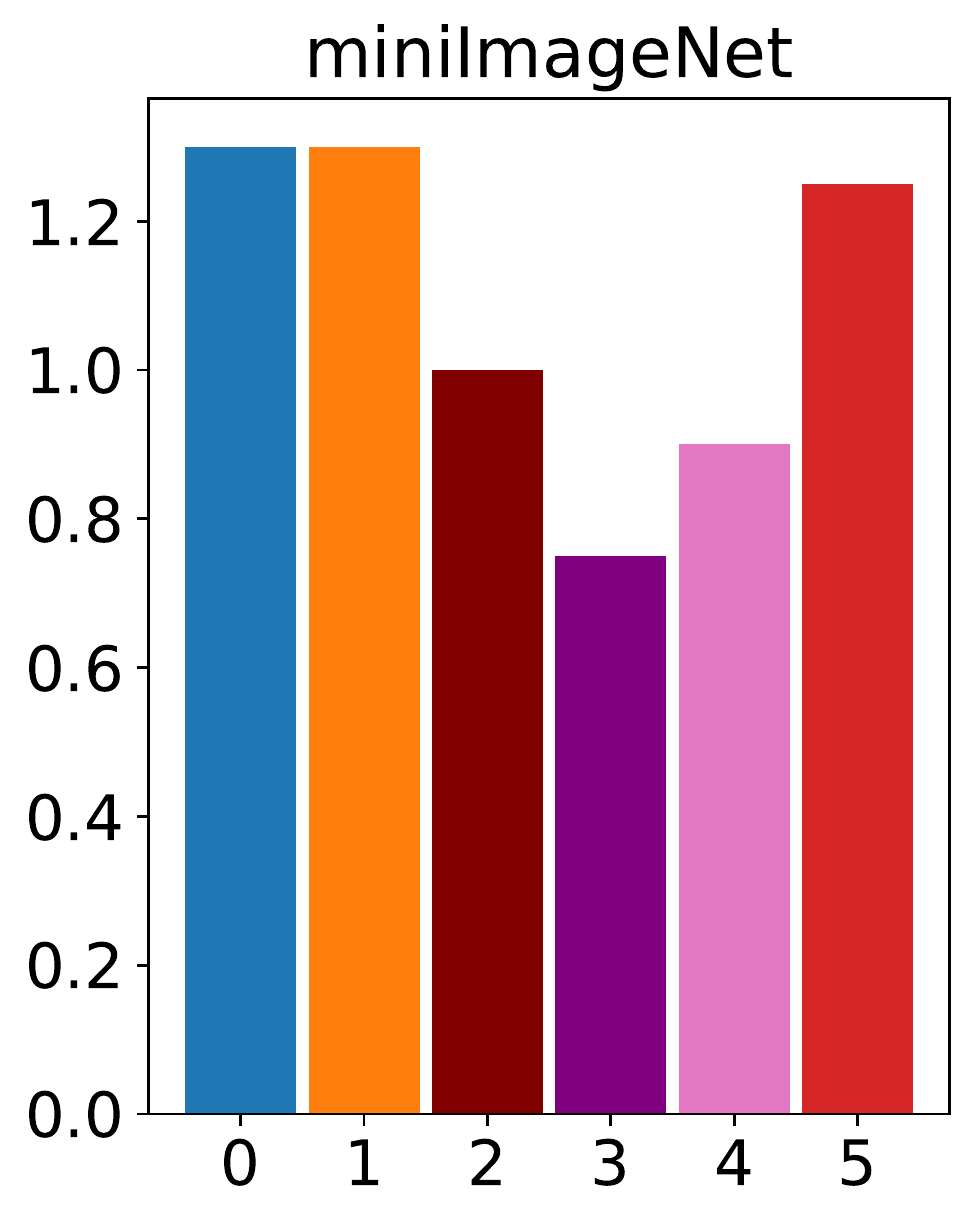}
    \includegraphics[width=0.21\linewidth]{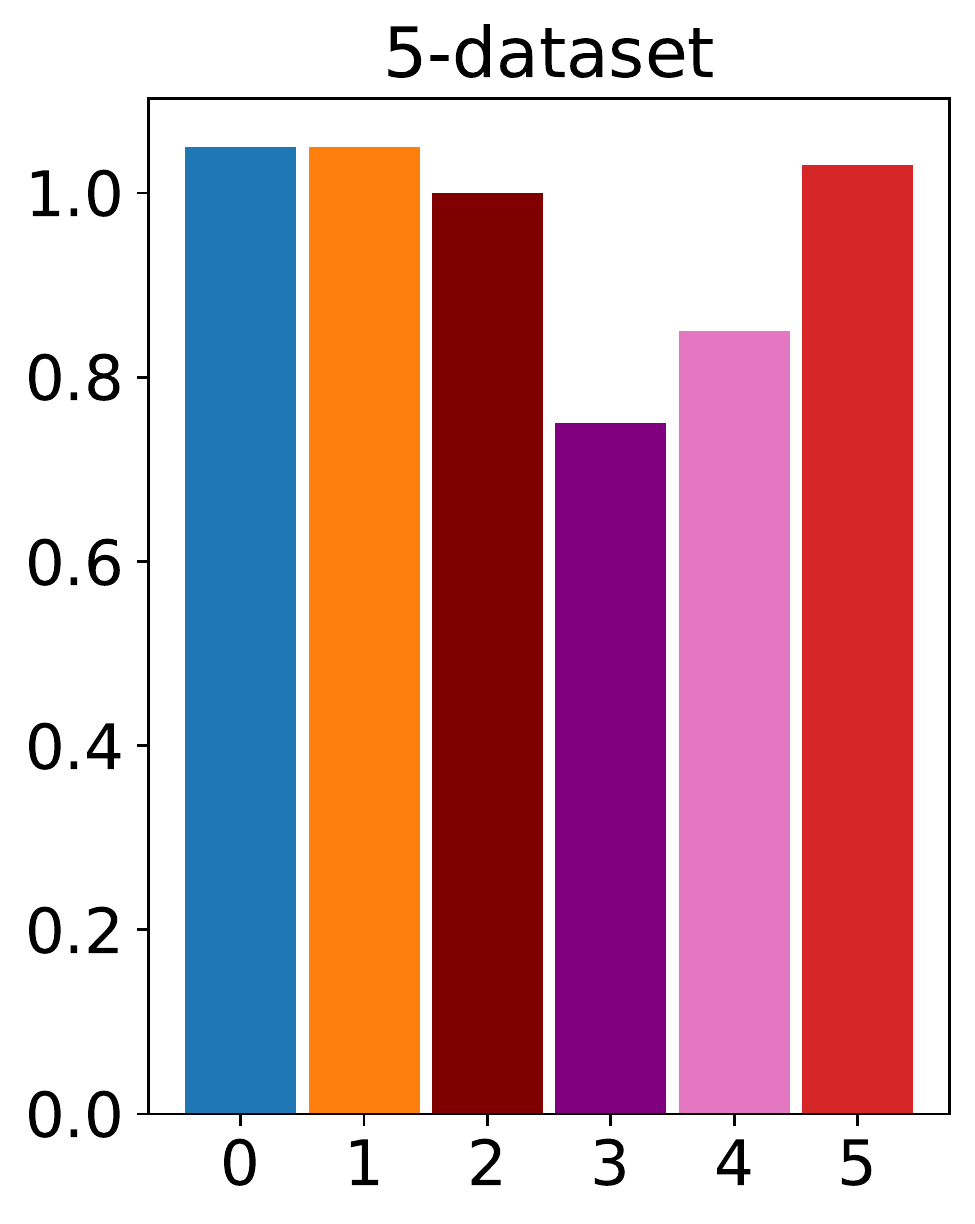}
    \includegraphics[width=0.1\linewidth]{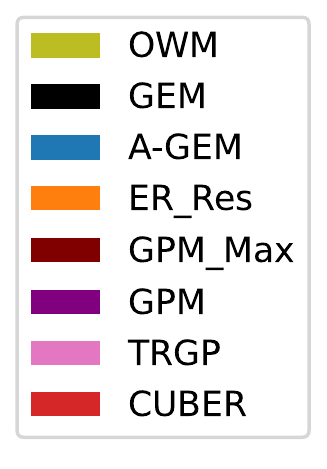}
    \caption{Comparison of memory utilization on PMNIST, CIFAR-100 Split, MiniImageNet and 5-Datasets, respectively.}
    \label{fig:mem}
\end{figure*}

\textbf{Memory and training time:} As shown in Figure \ref{fig:mem} and Table \ref{tab:time}, we compare the memory utilization and training time between CUBER and the related baseline methods in terms of the normalized value with respect to the value of GPM, following the same strategy as in \cite{lin2022trgp}. The values of OWM, EWC, HAT, A-GEM, ER$\_$Res are from the reported results in GPM \cite{saha2021gradient}. The value of TRGP is from the reported results in \cite{lin2022trgp}. It can be seen that CUBER indeed has comparable complexity with the baseline methods. While SVD is used to extract bases in CUBER, the training time of CUBER is still less than some baseline methods due to its relatively simple and fast model update. And because we only store part of the gradient for each old task, the required memory utilization does not increase  a lot.

\begin{table*}[!htbp]
\small
\centering
\caption{Training time comparison on CIFAR-100 Split, 5-Datasets and MiniImageNet. Here the training time is normalized with respect to the value of GPM. Please refer \cite{saha2021gradient} for more specific time.}
\label{tab:time}
\begin{tabular}{ccccccccc}
\toprule
\multirow{2}{*}{Dataset}          & \multicolumn{8}{c}{Methods}               \\ \cmidrule{2-9} 
                                 & OWM  & EWC & HAT & A-GEM & ER\_Res & GPM & TRGP & Ours (CUBER) \\ \midrule
\multicolumn{1}{l}{CIFAR-100} & 2.41 & 1.76 & 1.62 & 3.48 & 1.49 & 1 & 1.65 & 1.86\\
\multicolumn{1}{l}{5-Datasets} & - & 1.52 & 1.47 & 2.41 & 1.40 & 1 & 1.21 & 1.55 \\
\multicolumn{1}{l}{MiniImageNet} & - & 1.22 & 0.91 & 1.79 & 0.82 & 1 & 1.34 & 1.61 \\\bottomrule
\end{tabular}
\end{table*}

\subsection{Baseline implementations}

To ensure fair comparisons, we implement CUBER based on the released code of GPM \cite{saha2021gradient} and TRGP \cite{lin2022trgp}, and follow the exactly same experimental setups. We reproduced the reported results of GPM and TRGP using the provided hyperparameters in the papers. 
Therefore, for  other baseline methods considered in this work, we directly follow these two papers and present the reported results therein in Section 5. Besides, we also implemented all the baseline methods based on their released official code, and show the reproduced results in Table \ref{tab:repro}.

\begin{table*}[!htbp]
\scriptsize
\centering
\vspace{-0.2cm}
\caption{The ACC and BWT with the standard deviation values over 5 different runs on different datasets. Here for Split CIFAR-100, Split MiniImageNet and 5-Dataset we use a multi-head network, while we consider domain-incremental setup for Permuted MNIST. Moreover, $\epsilon_2=0.0$.}
\vspace{-0.1cm}
\label{tab:repro}
\scalebox{0.81}{
\begin{tabular}{c|cccccc|cc}
\toprule
\multicolumn{1}{c|}{\multirow{3}{*}{Method}} & \multicolumn{6}{c|}{Multi-head} & \multicolumn{2}{c}{Domain-incremental}\\\cmidrule{2-9}
& \multicolumn{2}{c}{Split CIFAR-100} &   \multicolumn{2}{c}{Split MiniImageNet} &   \multicolumn{2}{c|}{5-Dataset} &   \multicolumn{2}{c}{Permuted MNIST}\\ \cmidrule{2-9}
 & ACC(\%) & BWT(\%)   &   ACC(\%) & BWT(\%)   & ACC(\%) & BWT(\%)  & ACC(\%) & BWT(\%)   \\ \midrule
Multitask            & $79.58\pm 0.54$   & -     &   $69.46\pm 0.62$    & -     &   $91.54\pm 0.28$   & -&    $96.70\pm 0.02$  & -     \\ \midrule
OWM                  & $50.44\pm 0.72$   & $-30\pm 1$   & -   & -  & -       & -    &$89.63\pm 0.21$   & $-1\pm 0$   \\
EWC                  & $68.30\pm 0.65$   & $-2\pm 1$   & $50.78\pm 2.98$   & $-12\pm 4$ &   $87.67\pm 0.37$   & $-3\pm 1$ &    $89.05\pm 0.57$   & $-4\pm 2$ \\
HAT                  & $73.21\pm 0.76$   & $0\pm 0$     & $60.23\pm 0.57$   & $-4\pm 1$  & $91.82 \pm 0.34$  & $-1\pm 0$  & -   & -\\
A-GEM                & $64.16\pm 1.41$   & $-14\pm 3$   & $56.88\pm 0.87$   & $-13\pm 2$   & $82.48\pm 0.56$   & $-13\pm 1$   &  $83.05 \pm 0.12$   & $-14\pm 2$\\
ER\_Res              & $72.24\pm 0.57$   & $-7\pm 2$   & $58.04\pm 0.73$   & $-7\pm 1$   & $88.54\pm 0.36$   & $-4\pm 1$   &  $88.72\pm  0.56$  & $-10\pm 1$\\
GPM                  & $72.48\pm 0.40$   & $-0.9 \pm 0$  & $60.41\pm 0.61$   & $-0.7\pm 0.4$   & $91.22\pm 0.20$   &   $-1\pm 0 $   &  $93.91\pm 0.16$   & $-3\pm 0$ \\
TRGP & $74.46\pm 0.32$ & $-0.9\pm 0.01$  &$61.78\pm 0.60$ &  $-0.5\pm0.6 $  & $\pmb{93.56}\pm 0.10$ & $-0.04\pm 0.01$  &  $96.34\pm 0.11$ & $-0.8\pm 0.1$
\\ \midrule
CUBER (ours)                 &
$\pmb{75.54} \pm 0.22$ & $\pmb{0.13} \pm 0.08$  &    
$\pmb{62.67} \pm 0.74$ & $\pmb{0.23} \pm 0.15$  &    
$\pmb{93.48} \pm 0.10 $& $\pmb{0.00} \pm 0.02 $ &      
  $\pmb{97.25} \pm 0.00$ & $\pmb{-0.02} \pm 0.00$  \\ \bottomrule
\end{tabular}
}
\end{table*}
\end{document}